\documentclass[pdflatex]{sn-jnl}

\usepackage[T1]{fontenc}
\usepackage{lmodern}
\usepackage{graphicx}%
\usepackage{multirow}%
\usepackage{amsmath,amssymb,amsfonts}%
\usepackage{amsthm}%
\usepackage{mathrsfs}%
\usepackage[title]{appendix}%
\usepackage{xcolor}%
\usepackage{textcomp}%
\usepackage{manyfoot}%
\usepackage{booktabs}%
\usepackage{algorithm}%
\usepackage{algorithmicx}%
\usepackage{algpseudocode}%
\usepackage{listings}%
\usepackage{subcaption}%
\usepackage{placeins}%
\usepackage{float} %

\theoremstyle{thmstyleone}%
\theoremstyle{thmstyletwo}%

\theoremstyle{thmstylethree}%

\begin{document}

\title[Article Title]{Experimental Comparison of Light-Weight and Deep CNN Models Across Diverse Datasets}



\author[1]{\fnm{Md. Hefzul Hossain} \sur{Papon}}\email{mdhefzulpapon@gmail.com}
\equalcont{These authors contributed equally to this work.}

\author[1]{\fnm{Shadman} \sur{Rabby}}\email{shadmanrabby.cse@gmail.com}
\equalcont{These authors contributed equally to this work.}


\affil[1]{\orgdiv{Department of Computer Science and Engineering}, \orgname{University of Dhaka}, \state{Dhaka}, \country{Bangladesh}}



\abstract{

Our results reveal that a well-regularized shallow architecture can serve as a highly competitive baseline across heterogeneous domains—from smart-city surveillance to agricultural variety classification—without requiring large GPUs or specialized pre-trained models. This work establishes a unified, reproducible benchmark for multiple Bangladeshi vision datasets and highlights the practical value of lightweight CNNs for real-world deployment in low-resource settings. 

}




\maketitle

\section{Introduction}

Deep learning--based image classification has become a core component of
modern computer vision systems, enabling a wide range of applications
including intelligent transportation, agricultural monitoring, and
urban infrastructure management. Traditional machine learning
approaches, such as Support Vector Machines (SVMs), Random Forests, and
handcrafted feature descriptors (e.g., SIFT and HOG), depend heavily on
manual feature engineering and often struggle to generalize across
diverse imaging conditions, complex backgrounds, and large-scale visual
variability. The advent of Convolutional Neural Networks (CNNs)
addressed these limitations by enabling end-to-end feature learning
directly from raw image data, resulting in substantial improvements in
accuracy and robustness across many vision tasks.

Over the past decade, CNN architectures have evolved significantly, with
design principles such as increased depth, residual connections, and
compound scaling playing a critical role in performance. Models such as
ResNet~\cite{he2016deep} and EfficientNet~\cite{tan2019efficientnet}
demonstrate how architectural innovations can improve representation
capacity and optimization stability. However, these performance gains
often come at the cost of increased parameter counts, memory
requirements, and computational complexity, which can limit their
practical applicability in resource-constrained environments.

This challenge is particularly relevant for real-world deployments in
developing regions, where applications in smart cities, traffic
monitoring, and agriculture frequently operate under hardware,
bandwidth, and energy constraints. Moreover, domain-specific datasets
often exhibit heterogeneous visual characteristics, including varying
illumination, cluttered backgrounds, occlusions, and fine-grained
inter-class differences. These conditions raise important questions
about whether high-capacity pretrained models are always necessary, or
whether carefully designed lightweight architectures can provide
competitive performance with significantly lower computational cost.

In this context, custom-designed CNN architectures offer an attractive
alternative. By explicitly controlling architectural depth, filter
progression, and receptive field growth, lightweight CNNs can be
tailored to domain-specific tasks without excessive overparameterization.
Recent datasets originating from the Bangladeshi context further
highlight this need, including road damage and manhole detection for
urban infrastructure monitoring~\cite{hossen2025road}, footpath
encroachment analysis for smart city management~\cite{lubaina2025footpathvision},
unauthorized vehicle detection for traffic surveillance~\cite{das2025detecting},
and agricultural image classification tasks such as MangoImageBD and
PaddyVarietyBD~\cite{ferdaus2025mangoimagebd,tahsin2025paddy}. Together,
these datasets span diverse visual domains and classification
difficulties, making them well-suited for evaluating the generalization
capability of CNN architectures.

Motivated by these considerations, this study conducts a comparative
evaluation of a custom-designed lightweight CNN against widely adopted
deep learning models, namely EfficientNetB0 and ResNet18, trained both
from scratch and using transfer learning. The custom model is designed
from first principles with a focus on computational efficiency, stable
training behavior, and deployment feasibility. Through systematic
experimentation across five publicly available datasets, this work
addresses the following research questions:

\begin{enumerate}
    \item Can a lightweight, custom CNN achieve competitive performance
    compared to standard deep CNN architectures across diverse visual
    domains?
    \item How do architectural complexity and training strategy
    (from-scratch versus transfer learning) influence performance and
    generalization on domain-specific datasets?
    \item What trade-offs emerge between classification accuracy,
    computational efficiency, and deployment suitability in
    resource-limited environments?
\end{enumerate}

The main contributions of this work are summarized as follows:
\begin{itemize}
    \item[(1)] The design and evaluation of a compact CustomCNN tailored
    for efficient image classification under resource constraints.
    \item[(2)] A unified experimental benchmark across five diverse
    datasets spanning transportation, urban infrastructure, and
    agriculture.
    \item[(3)] A detailed comparative analysis highlighting performance,
    efficiency, and deployment trade-offs between lightweight and
    standard deep CNN architectures.
\end{itemize}

\section{Proposed Method}
\label{sec:method}

In this section, we present our proposed end-to-end image classification framework based on a compact Convolutional Neural Network (CNN). The entire pipeline consists of four components: (i) dataset construction and stratified splitting, (ii) a lightweight CNN architecture with progressive feature extraction, (iii) an efficient training strategy with regularization and learning-rate scheduling, and (iv) performance stabilization using early stopping. 

\subsection{Dataset Construction and Preprocessing}

Each dataset is organized in a hierarchical directory structure, where each top-level folder corresponds to a unique class label. All images within each class directory, including nested subfolders, are recursively scanned and indexed. Let
\[
\mathcal{D} = \{(x_i, y_i)\}_{i=1}^{N}
\]
denote the complete dataset, where $x_i$ represents an input image and $y_i \in \{1, \dots, C\}$ denotes the associated class index.

To ensure a fair and balanced evaluation protocol, a \emph{stratified dataset splitting strategy} is adopted. For each class, samples are randomly shuffled and divided into training, validation, and test subsets using a ratio of $70\%$, $15\%$, and $15\%$, respectively. This strategy preserves the original class distribution across all splits and prevents bias toward majority classes, even in datasets exhibiting class imbalance.

In addition to stratified splitting, class imbalance is explicitly addressed during training through the use of a weighted loss function. Class weights are computed inversely proportional to class frequencies in the training set and, when enabled, are incorporated into the cross-entropy loss. This weighting mechanism penalizes misclassification of under-represented classes more strongly, thereby encouraging balanced learning without altering the original data distribution through over- or under-sampling.

All images are resized to $224 \times 224$ pixels and normalized using ImageNet mean and standard deviation statistics to ensure compatibility with pretrained architectures. Data augmentation is optionally applied to the training set and includes random horizontal flipping, random rotation within $\pm 10^\circ$, and color jittering of brightness, contrast, and saturation. These augmentations introduce controlled appearance variability and help improve model generalization while maintaining label consistency.

\subsection{Custom CNN Architecture}

\begin{figure}[!ht]
    \centering
    \includegraphics[width=\textwidth]{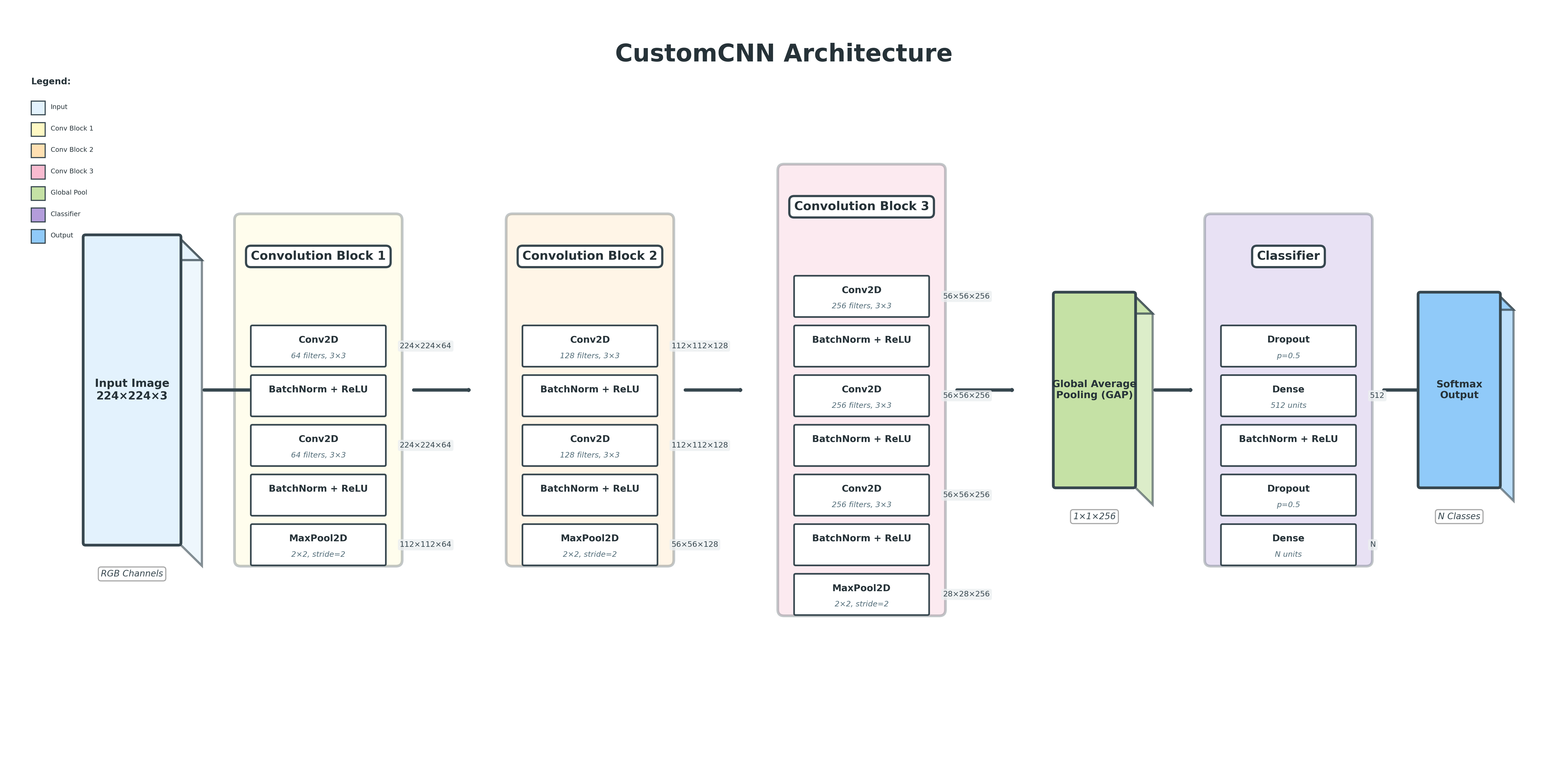}
    \caption{Architecture diagram of the proposed Custom CNN model.}
    \label{fig:custom_cnn_arch}
\end{figure}

We propose a compact yet expressive CNN architecture designed for efficient feature extraction while maintaining a low computational footprint. The model follows a hierarchical structure comprising three convolutional blocks, a global pooling layer, and a fully connected classifier. Overall architecture of the model is illustrated in Figure ~\ref{fig:custom_cnn_arch}.

\subsubsection{Convolutional Feature Extractor}

The feature extractor $\mathcal{F}$ is composed of three convolutional blocks:
\[
\mathcal{F} = [B_1, B_2, B_3],
\]
where each block $B_k$ contains multiple convolutional layers followed by batch normalization, ReLU activation, and a downsampling max-pooling operation.

\paragraph{Block Definitions.}  
Each block $B_k$ is constructed as:
\[
\text{Conv}_{3\times 3} \rightarrow \text{BatchNorm} \rightarrow \text{ReLU},
\]
repeated $n_k$ times (with $n_k = \{2, 2, 3\}$ for the three blocks). Pooling reduces the spatial resolution by a factor of 2.

The channel progression is:
\[
3 \rightarrow 64 \rightarrow 128 \rightarrow 256,
\]
allowing deeper layers to capture progressively high-level semantic information.

\subsubsection{Global Average Pooling and Classifier}

After the final convolutional block, we apply global average pooling (GAP) to reduce the spatial dimension to $1 \times 1$, yielding a compact descriptor of dimension 256:
\[
z = \text{GAP}(B_3(x)).
\]

The classifier consists of two fully connected layers with dropout and batch normalization to reduce overfitting:
\[
z \rightarrow \text{FC}_{256\rightarrow 512} \rightarrow \text{BatchNorm} \rightarrow \text{ReLU} \rightarrow \text{Dropout} \rightarrow \text{FC}_{512\rightarrow C}.
\]

The final output logits are passed to a softmax layer during inference for class probability estimation.

\subsubsection{Weight Initialization}
To stabilize training, we adopt He initialization for convolutional layers:
\[
W \sim \mathcal{N}(0, \sqrt{2/\text{fan\_out}}),
\]
and small Gaussian initialization for linear layers:
\[
W \sim \mathcal{N}(0, 0.01).
\]
Batch normalization parameters are initialized with $\gamma = 1$ and $\beta = 0$.

\begin{algorithm}[htbp]
\caption{Training Pipeline for the Proposed CNN Model}
\label{alg:training}
\begin{algorithmic}[1]
\State Load dataset and perform stratified splitting.
\State Apply preprocessing and data augmentation.
\State Initialize CNN model parameters.
\For{epoch $= 1$ to $E$}
    \State Train model on minibatches from training set.
    \State Evaluate on validation set.
    \State Update learning rate using scheduler.
    \If{early stopping criterion satisfied}
        \State \textbf{break}
    \EndIf
\EndFor
\State Evaluate final model on the test set.
\end{algorithmic}
\end{algorithm}

\subsection{Training Strategy}

The model is trained using the Adam optimizer with learning rate $\eta = 10^{-3}$ (configurable), and weight decay of $10^{-4}$ for regularization. The loss function is the standard cross-entropy loss:
\[
\mathcal{L} = - \sum_{i=1}^{C} y_i \log(\hat{y}_i).
\]

During training, we employ the ReduceLROnPlateau learning rate scheduler. The learning rate is halved whenever the validation loss fails to improve for 5 consecutive epochs.

\subsection{Early Stopping}

To avoid overfitting and unnecessary computation, an early stopping criterion is incorporated. Training terminates if validation loss does not show an improvement of at least $\delta = 0.001$ for $p = 10$ epochs:
\[
\text{stop if} \quad \mathcal{L}_{val}^{(t)} > \mathcal{L}_{val}^{*} - \delta.
\]

\subsection{Overall Workflow}

Algorithm~\ref{alg:training} summarizes the complete training pipeline.

The combination of careful data preparation, architectural compactness, and robust training strategies ensures strong generalization performance while keeping the computational overhead modest.

\section{Experimental Results}

This section presents a comprehensive evaluation of the proposed
lightweight CustomCNN across five recent Bangladeshi vision datasets,
namely RoadDamageBD~\cite{hossen2025road}, PaddyVarietyBD~\cite{tahsin2025paddy},
MangoImageBD~\cite{ferdaus2025mangoimagebd}, FootpathVision~\cite{lubaina2025footpathvision},
and Auto-RickshawImageBD~\cite{das2025detecting}. The analysis is
structured in two stages. First, we provide a detailed, dataset-wise
examination of the performance and learning behavior of the CustomCNN.
Subsequently, we present a comparative evaluation against standard deep
CNN architectures trained both from scratch and using transfer learning.

The proposed CustomCNN consists of three convolutional blocks with
progressively increasing channel widths (64–128–256), followed by
global average pooling and a two-layer fully connected classification
head. The architecture is intentionally designed to remain compact and
computationally efficient, enabling end-to-end training on modest GPU
resources while serving as a unified baseline across diverse visual
domains. Unlike deeper architectures such as ResNet18 or EfficientNetB0,
which rely on higher model capacity or pretrained representations, the
CustomCNN emphasizes simplicity, stable optimization, and deployment
feasibility.

All experiments follow a consistent training protocol to ensure fair
comparison. Input images are resized to $224 \times 224$ pixels,
normalized using ImageNet statistics, and augmented during training with
random horizontal flips, small rotations, and mild color jitter. The
models are trained using the Adam optimizer with cross-entropy loss,
along with a ReduceLROnPlateau learning rate scheduler and early
stopping based on validation loss. No dataset-specific architectural or
hyperparameter tuning is applied, allowing performance differences to
reflect model capacity and training strategy rather than manual
optimization.

    
    
    
    
    
    

\subsection*{Overall Quantitative Performance}

Table~\ref{tab:overall-results} reports the classification performance
of the proposed CustomCNN across five datasets using the final test
results obtained from the experimental logs. The evaluation spans
diverse visual domains, including agricultural imagery and complex
urban scenes, while employing the same compact backbone and training
configuration throughout.

Across all datasets, the CustomCNN achieves test accuracies ranging
from $72.00\%$ to $94.12\%$. The strongest performance is observed on
RoadDamageBD, where the model attains a test accuracy of $94.12\%$ with
a macro F1-score of $92.44\%$, indicating robust generalization to
unseen road damage patterns. MangoImageBD also demonstrates strong
performance, with a test accuracy of $90.65\%$, confirming that the
network effectively captures salient visual cues such as color and
texture in agricultural imagery.

Performance on PaddyVarietyBD is comparatively lower, with a test
accuracy of $78.76\%$, reflecting the fine-grained nature of the task
and the high visual similarity among classes. For FootpathVision, the
model achieves a test accuracy of $86.02\%$, while Auto-RickshawImageBD
records a test accuracy of $72.00\%$. These datasets contain cluttered
street scenes, occlusions, and varying viewpoints, which naturally
increase classification difficulty. Nevertheless, the CustomCNN
maintains stable and competitive performance without any
dataset-specific architectural adjustments.

\subsection*{Per-dataset Learning Behaviour}
Table~\ref{tab:overall-results} summarizes the final test performance
obtained on each dataset using the proposed CustomCNN. The reported
metrics are computed on the held-out test split and are extracted
directly from the experimental logs. Despite the diversity of visual
content—ranging from fine-grained agricultural imagery to complex urban
street scenes—the same compact backbone demonstrates consistently
competitive performance without dataset-specific architectural
adaptations.
\begin{table}[t]
    \centering
    \caption{Final test performance of the proposed CustomCNN on the
    five datasets. All metrics are reported on the held-out test split.}
    \label{tab:overall-results}
    \begin{tabular}{lcccc}
        \hline
        \textbf{Dataset} 
        & \textbf{Test Acc. (\%)} 
        & \textbf{Prec. (\%)} 
        & \textbf{Recall (\%)} 
        & \textbf{F1 (\%)} \\
        \hline
        RoadDamageBD~\cite{hossen2025road}
        & 94.12 & 94.12 & 91.08 & 92.44 \\

        PaddyVarietyBD~\cite{tahsin2025paddy}
        & 78.76 & 80.27 & 78.76 & 78.29 \\

        MangoImageBD~\cite{ferdaus2025mangoimagebd}
        & 90.65 & 82.88 & 84.51 & 82.07 \\

        FootpathVision~\cite{lubaina2025footpathvision}
        & 86.02 & 86.04 & 85.11 & 85.48 \\

        Auto-RickshawImageBD~\cite{das2025detecting}
        & 72.00 & 67.60 & 73.59 & 67.83 \\
        \hline
    \end{tabular}
\end{table}

\paragraph{RoadDamageBD}
On RoadDamageBD~\cite{hossen2025road}, the CustomCNN achieves a test
accuracy of $94.12\%$ with a macro F1-score of $92.44\%$. The smooth loss
decay and strong agreement between precision and recall indicate stable
optimization and effective feature learning for road surface damage and
manhole patterns.

\paragraph{PaddyVarietyBD.}
PaddyVarietyBD~\cite{tahsin2025paddy} represents a challenging
fine-grained classification problem due to subtle inter-class visual
differences. The CustomCNN reaches a test accuracy of $78.76\%$ and a
macro F1-score of $78.29\%$. These results indicate that, despite its
compact design, the network is able to learn moderately discriminative
features for visually similar paddy varieties.

\paragraph{MangoImageBD.}
For MangoImageBD~\cite{ferdaus2025mangoimagebd}, the model achieves a
test accuracy of $90.65\%$ with a macro F1-score of $82.07\%$. The close
alignment between precision and recall suggests balanced classification
behavior, demonstrating the suitability of the CustomCNN for
well-curated datasets with clear inter-class differences.

\paragraph{FootpathVision.}
On FootpathVision~\cite{lubaina2025footpathvision}, the CustomCNN
achieves a test accuracy of $86.02\%$ and a macro F1-score of $85.48\%$.
The strong performance indicates that the learned features generalize
effectively to complex urban footpath scenes containing background
clutter and surface irregularities.

\paragraph{Auto-RickshawImageBD.}
For the Auto-RickshawImageBD dataset~\cite{das2025detecting}, the model
records a test accuracy of $72.00\%$ and a macro F1-score of $67.83\%$.
Given the presence of heavy traffic, occlusions, and viewpoint
variations, these results demonstrate that the CustomCNN can still
extract meaningful visual representations in challenging smart-city
environments, albeit with reduced performance due to increased scene
complexity.

\subsection{Comparing the performance of the  Proposed Custom CNN Model with Pretrained models and Transfer learning models}

\textbf{RoadDamageBD}\\
The experimental results on the RoadDamageBD dataset, obtained from 68 evaluation samples (15\% of the test set), show that the custom-designed CNN provides an effective balance between classification performance and computational efficiency (see  Table~\ref{tab:performance_RoadDamageBD}). Despite its relatively lightweight architecture, the CustomCNN achieves an accuracy of 0.941 and a competitive F1-score of 0.924, while maintaining a significantly lower parameter count and memory footprint compared to deeper models. This indicates that a task-specific architecture can successfully capture the structural characteristics of road damage patterns in the RoadDamageBD dataset without requiring excessive model complexity.

In contrast, deep pretrained architectures trained from scratch perform noticeably worse on the RoadDamageBD dataset. EfficientNetB0 without transfer learning shows substantial degradation across all evaluation metrics, suggesting that the dataset size and variability are insufficient to effectively train such high-capacity models from random initialization. This highlights the limitations of purely data-driven training for large architectures in domain-specific datasets, where overparameterization can negatively impact generalization.

The transfer learning-based approaches demonstrate the strongest overall performance on the RoadDamageBD dataset, achieving the highest recall and F1-scores across models. EfficientNetB0 and ResNet18 with transfer learning leverage representations learned on large-scale datasets, leading to improved robustness and generalization. However, these gains come at the cost of increased training time, higher computational requirements, and larger memory footprints, particularly for ResNet18. Therefore, while transfer learning is advantageous when predictive performance is the primary objective, the CustomCNN remains a compelling choice for practical road damage detection applications where computational efficiency and deployment constraints are critical.\\
The learning curves show that the CustomCNN converges smoothly, with training and validation loss decreasing in a consistent manner and only minor divergence, indicating stable optimization and limited overfitting. EfficientNetB0 trained from scratch exhibits slower convergence and noticeable gaps between training and validation accuracy, suggesting insufficient data to effectively train a high-capacity model from random initialization. The transfer learning-based models demonstrate the fastest convergence and the most stable learning behavior. Both EfficientNetB0 and ResNet18 with transfer learning achieve rapid improvements in validation accuracy during early epochs, followed by gradual saturation, indicating efficient reuse of pretrained representations. While these models deliver superior final accuracy and F1-scores, their learning curves also reflect longer training durations and higher computational demands. Overall, the learning curve analysis reinforces the quantitative results, confirming that CustomCNN offers a favorable balance between convergence stability, generalization, and computational efficiency, whereas transfer learning provides performance gains at the cost of increased complexity.
\begin{table*}[t]
\caption{Performance comparison of different models on the RoadDamageBD Dataset}
\label{tab:performance_RoadDamageBD}
\centering
\small
\setlength{\tabcolsep}{4pt}
\resizebox{\textwidth}{!}{%
\begin{tabular}{l c c c c c c c}
\toprule
\textbf{Model} 
& \textbf{Acc.} 
& \textbf{Prec.} 
& \textbf{Recall} 
& \textbf{F1} 
& \textbf{Train Time (s)} 
& \textbf{Params} 
& \textbf{Size (MB)} \\
\midrule
CustomCNN 
& 0.941 & 0.941 & 0.911 & 0.924 
& 883.62 & 1,871,426 & 7.15 \\

EfficientNetB0 (Scratch) 
& 0.721 & 0.360 & 0.500 & 0.419 
& 559.56 & 4,010,110 & 15.46 \\

EfficientNetB0 (Transfer) 
& 0.941 & 0.918 & 0.943 & 0.929 
& 4631.07 & 4,010,110 & 15.46 \\

ResNet18 (Scratch) 
& 0.941 & 0.918 & 0.943 & 0.929 
& 1353.52 & 11,177,538 & 42.68 \\

ResNet18 (Transfer) 
& 0.926 & 0.904 & 0.917 & 0.910 
& 4656.91 & 11,177,538 & 42.68 \\
\bottomrule
\end{tabular}
}
\end{table*}




\begin{figure}[t]
\centering

\begin{subfigure}{0.49\columnwidth}
  \includegraphics[width=\linewidth]{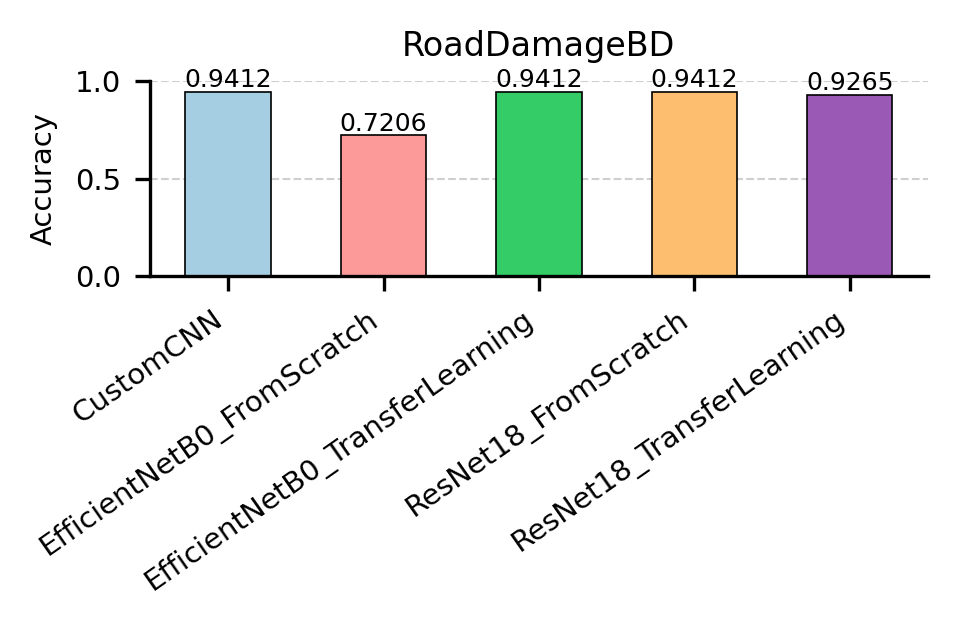}
  \caption{Accuracy}
\end{subfigure}
\hfill
\begin{subfigure}{0.49\columnwidth}
  \includegraphics[width=\linewidth]{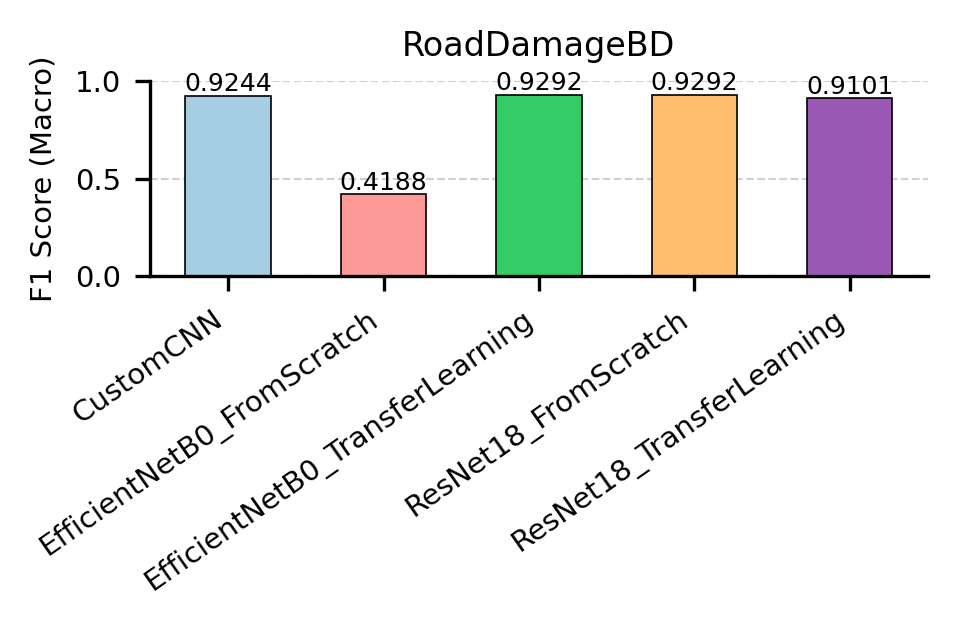}
  \caption{F1-score}
\end{subfigure}

\vspace{1.5mm}

\begin{subfigure}{0.49\columnwidth}
  \includegraphics[width=\linewidth]{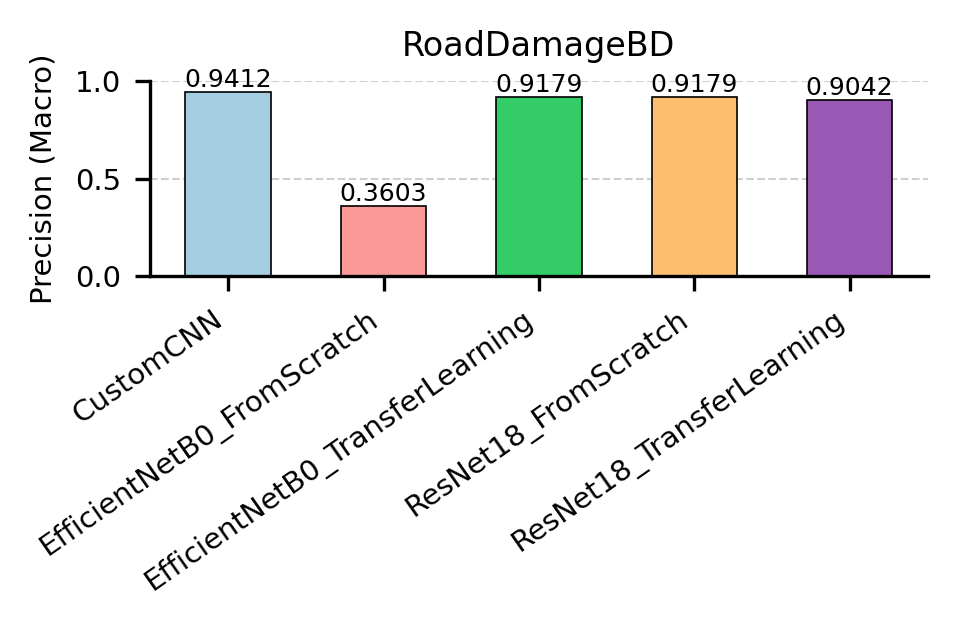}
  \caption{Precision}
\end{subfigure}
\hfill
\begin{subfigure}{0.49\columnwidth}
  \includegraphics[width=\linewidth]{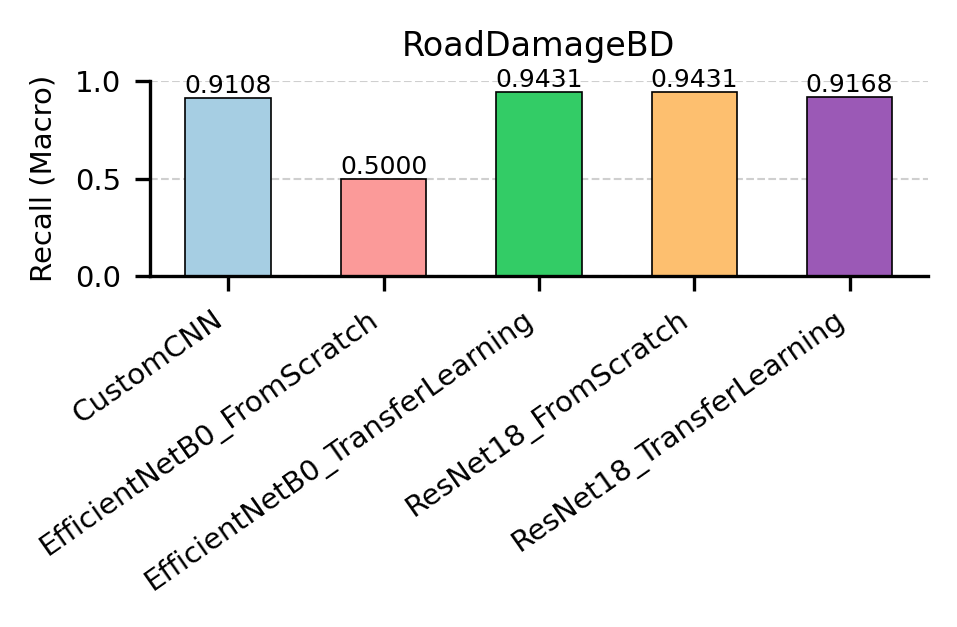}
  \caption{Recall}
\end{subfigure}

\vspace{1.5mm}

\begin{subfigure}{0.49\columnwidth}
  \includegraphics[width=\linewidth]{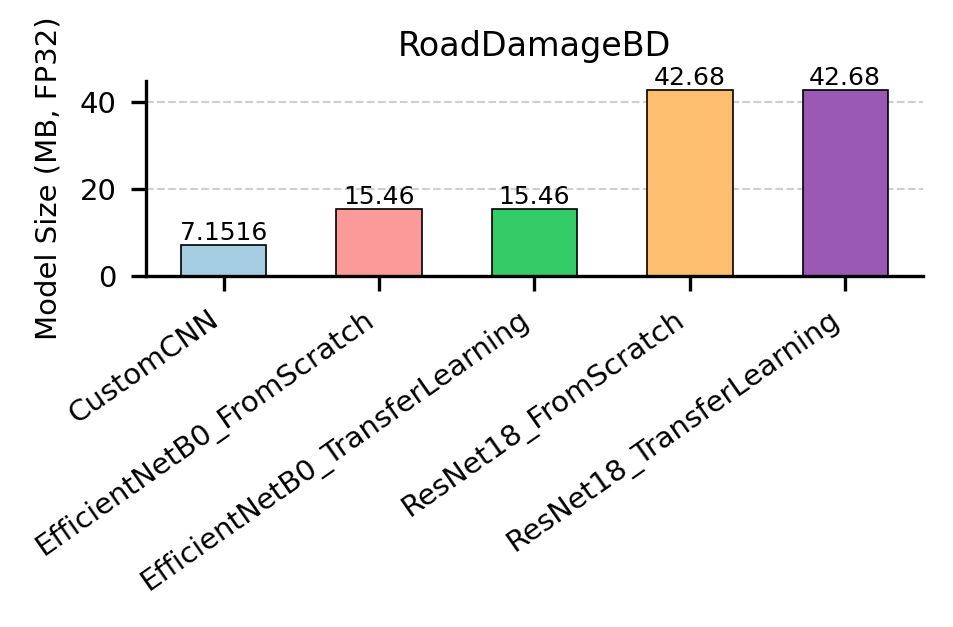}
  \caption{Model size}
\end{subfigure}
\hfill
\begin{subfigure}{0.49\columnwidth}
  \includegraphics[width=\linewidth]{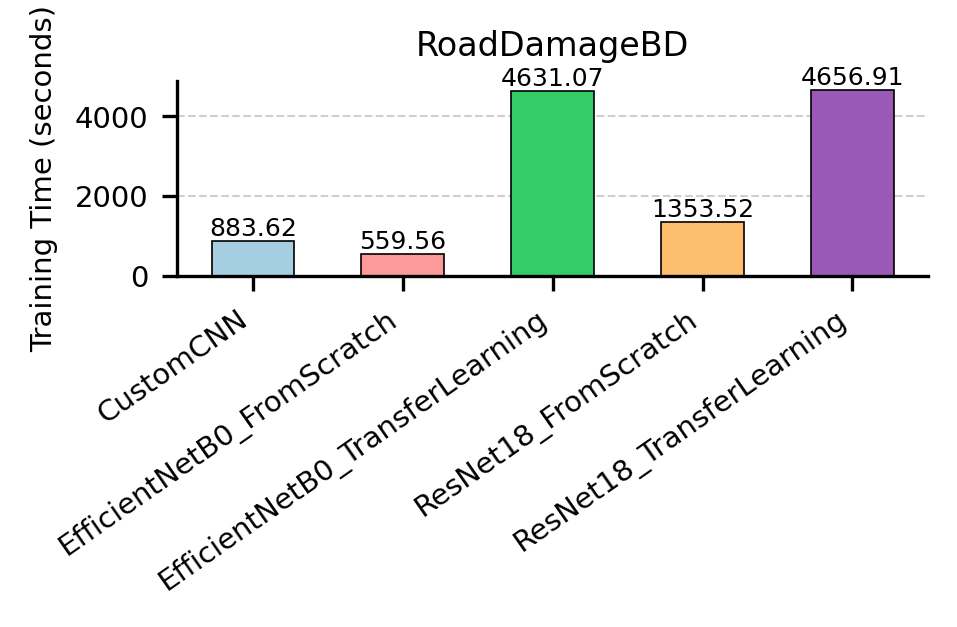}
  \caption{Training time}
\end{subfigure}

\caption{Performance comparison of CustomCNN, scratch-trained, and transfer learning models on the RoadDamageBD dataset.}
\label{fig:roadimagebd_performance}
\end{figure}

\FloatBarrier

\begin{figure}[htbp]
\centering
\includegraphics[width=0.80\textwidth]{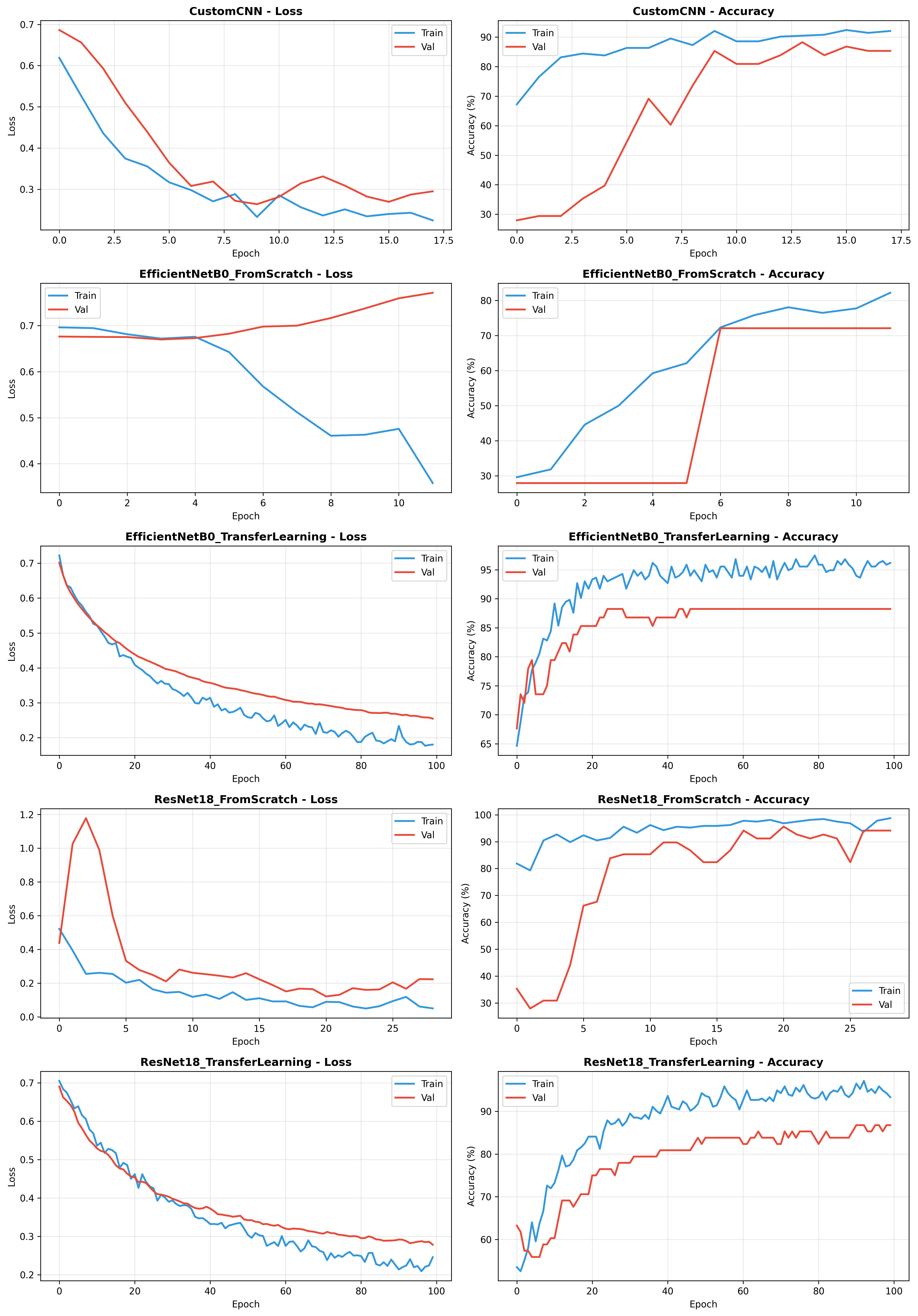} 
\caption{Comaprison of training Curve for testing models on RoadDamageBD dataset }
\label{fig:roadimagebd_training_curve}
\end{figure}

\textbf{Footpath Vision}\\
The experimental results on the FootpathVision dataset, as reported in Table~\ref{tab:performance_table_footpath}, demonstrate notable differences in performance and training behavior across custom-designed, scratch-trained, and transfer learning–based models. The CustomCNN achieves competitive baseline performance while maintaining a compact architecture and reduced computational cost. Although its accuracy and macro F1-score are lower than those of deeper networks, the relatively small model size and stable convergence make it a practical choice for real-time footpath analysis in resource-constrained environments.

Models trained from scratch exhibit varying degrees of effectiveness on the FootpathVision dataset. EfficientNetB0 trained without pretraining shows moderate improvements over the CustomCNN; however, its learning curves indicate slower convergence and higher validation loss fluctuations during early epochs. ResNet18 trained from scratch demonstrates stronger representational capacity, achieving higher classification performance, but the validation loss curves reveal intermittent spikes and corresponding drops in validation accuracy. This behavior suggests sensitivity to sample imbalance and local texture variations commonly found in footpath imagery, indicating a risk of overfitting when training deeper architectures from random initialization.

Transfer learning–based models show the most stable and consistent learning behavior on the FootpathVision dataset. Both EfficientNetB0 and ResNet18 with pretrained weights exhibit smooth loss decay and closely aligned training and validation accuracy curves throughout training. This indicates improved generalization and reduced training volatility, as pretrained features effectively capture low-level structural patterns such as cracks, surface irregularities, and texture discontinuities. While transfer learning leads to increased training time and larger memory footprints, particularly for ResNet18, it delivers superior robustness and reliability in classification performance.

The learning curves show that transfer learning models achieve rapid accuracy gains within the first few epochs, followed by smooth saturation and consistently lower validation loss, indicating stable feature reuse and effective generalization. In contrast, scratch-trained EfficientNetB0 and ResNet18 exhibit delayed validation accuracy improvements and noticeable loss oscillations, particularly in early epochs, reflecting unstable optimization. The CustomCNN demonstrates steady but slower convergence, with a relatively small gap between training and validation curves, suggesting controlled capacity and reduced overfitting risk. These trends confirm that while scratch-trained deep models can eventually converge, transfer learning provides more reliable and stable training behavior on the FootpathVision dataset.
\begin{table*}[t]
\caption{Performance comparison of different models on the Footpath Vision dataset}
\label{tab:performance_table_footpath}
\centering
\small
\setlength{\tabcolsep}{4pt}
\resizebox{\textwidth}{!}{%
\begin{tabular}{l c c c c c c c}
\toprule
\textbf{Model} 
& \textbf{Acc.} 
& \textbf{Prec.} 
& \textbf{Recall} 
& \textbf{F1} 
& \textbf{Train Time (s)} 
& \textbf{Params} 
& \textbf{Size (MB)} \\
\midrule
CustomCNN 
& 0.860 & 0.860 & 0.851 & 0.855 
& 1708.32 & 1,871,426 & 7.15 \\

EfficientNetB0 
& 0.812 & 0.807 & 0.811 & 0.808 
& 3570.80 & 4,010,110 & 15.46 \\

EfficientNetB0 (Transfer) 
& 0.871 & 0.869 & 0.866 & 0.867 
& 8940.65 & 4,010,110 & 15.46 \\

ResNet18 (Scratch) 
& 0.806 & 0.805 & 0.794 & 0.798 
& 1606.79 & 11,177,538 & 42.68 \\

ResNet18 (Transfer) 
& 0.823 & 0.818 & 0.822 & 0.819 
& 8162.96 & 11,177,538 & 42.68 \\
\bottomrule
\end{tabular}
}
\end{table*}

\begin{figure}[t]
\centering

\begin{subfigure}{0.49\columnwidth}
  \includegraphics[width=\linewidth]{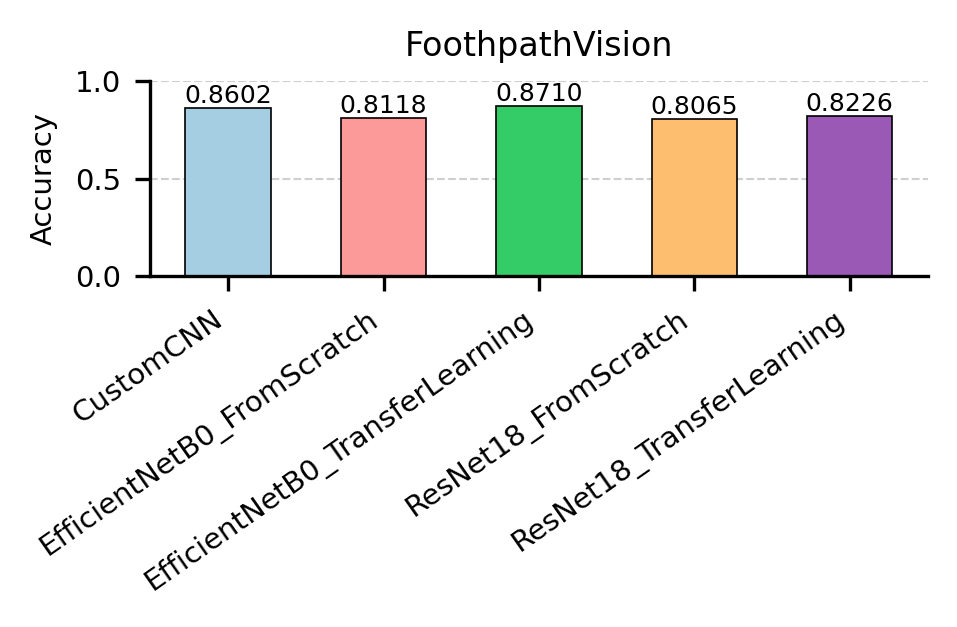}
  \caption{Accuracy}
\end{subfigure}
\hfill
\begin{subfigure}{0.49\columnwidth}
  \includegraphics[width=\linewidth]{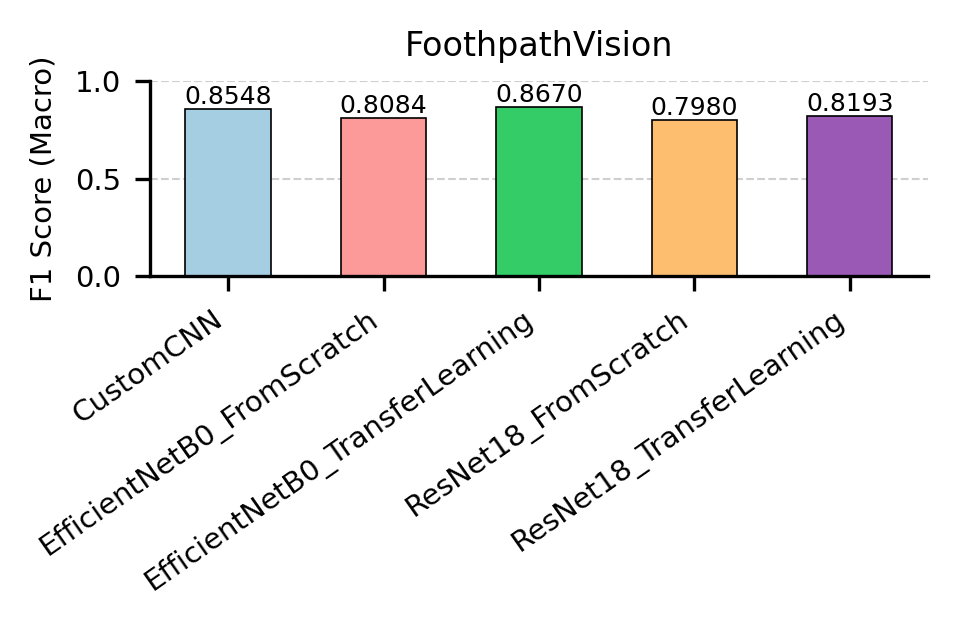}
  \caption{F1-score}
\end{subfigure}

\vspace{1.5mm}

\begin{subfigure}{0.49\columnwidth}
  \includegraphics[width=\linewidth]{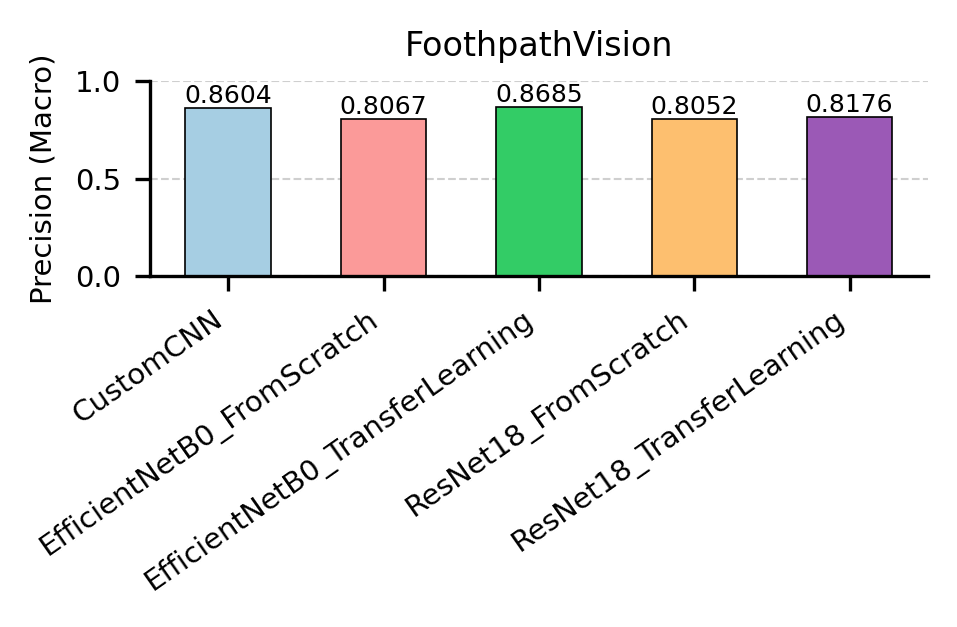}
  \caption{Precision}
\end{subfigure}
\hfill
\begin{subfigure}{0.49\columnwidth}
  \includegraphics[width=\linewidth]{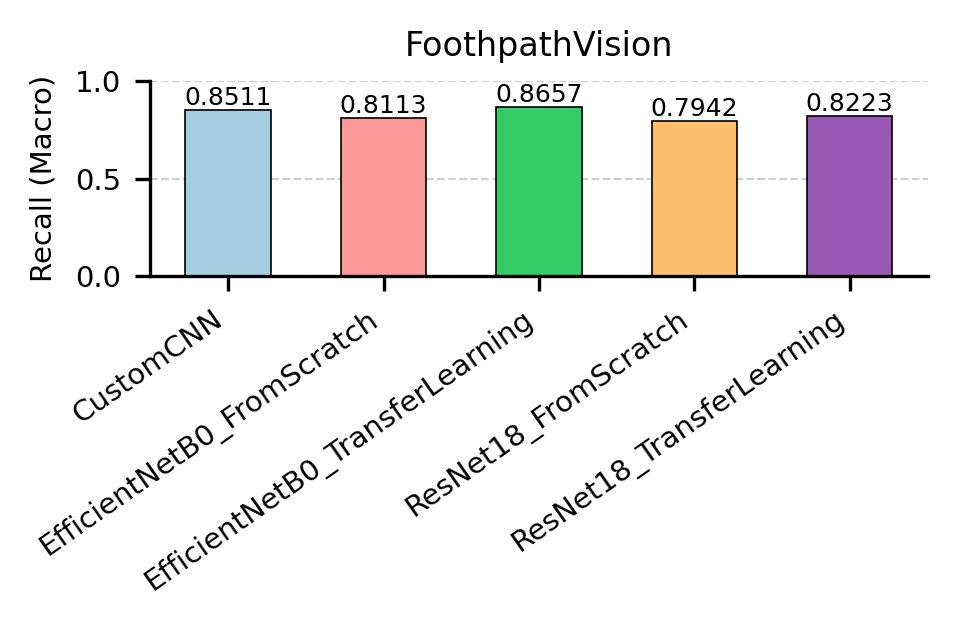}
  \caption{Recall}
\end{subfigure}

\vspace{1.5mm}

\begin{subfigure}{0.49\columnwidth}
  \includegraphics[width=\linewidth]{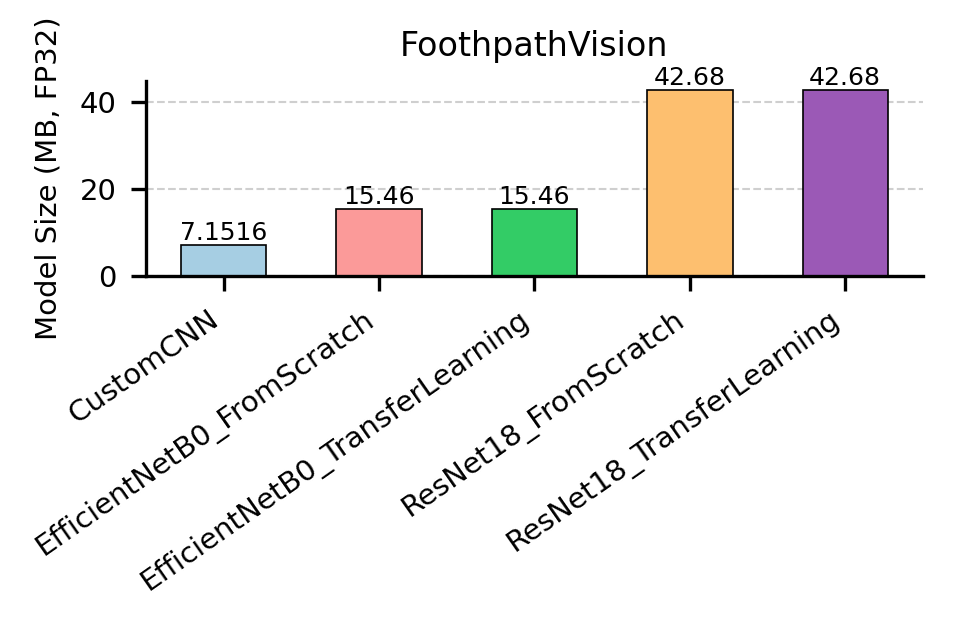}
  \caption{Model size}
\end{subfigure}
\hfill
\begin{subfigure}{0.49\columnwidth}
  \includegraphics[width=\linewidth]{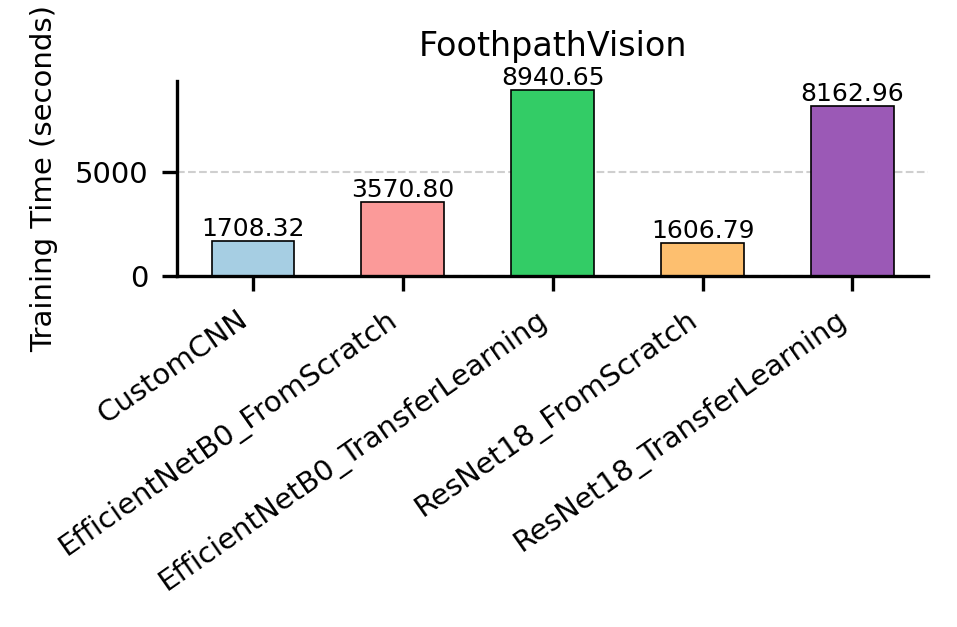}
  \caption{Training time}
\end{subfigure}

\caption{Performance comparison of CustomCNN, scratch-trained, and transfer learning models on the FootpathVision dataset.}
\label{fig:performance_compare_footpath}
\end{figure}

\FloatBarrier

\begin{figure}[htbp]
\centering
\includegraphics[width=0.80\textwidth]{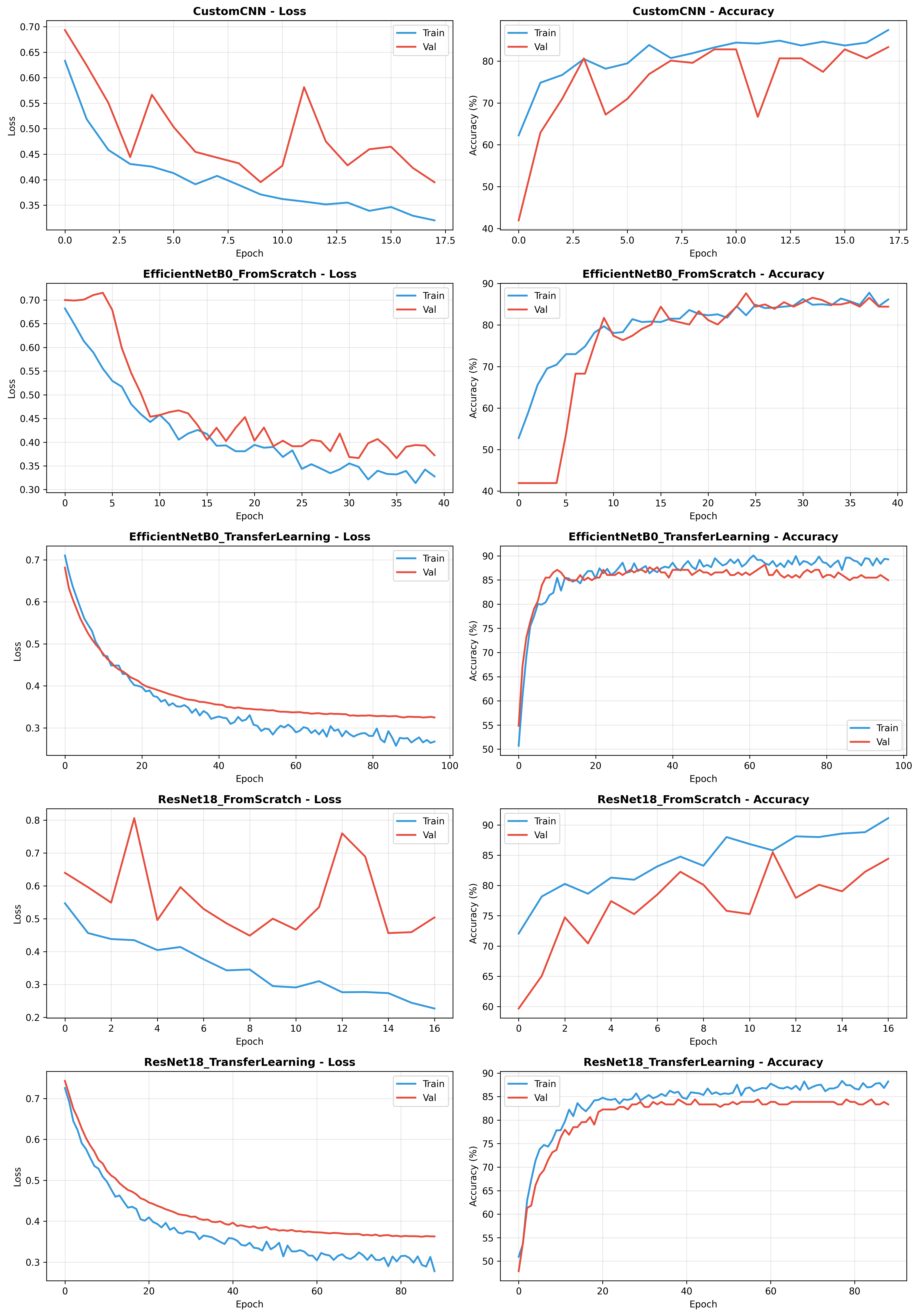} 
\caption{Comaprison of training Curve for testing models on Footpath Vision dataset }
\label{fig:training_curve_footpath}
\end{figure}

\textbf{MangoImage Dataset}\\
The experimental results on the MangoImageBd dataset, summarized in Table~\ref{tab:performance_table_MangoImageBd}, reveal distinct performance trends across custom-designed, scratch-trained, and transfer learning–based architectures. The CustomCNN establishes a strong baseline, achieving an accuracy of 0.907 and a macro F1-score of 0.821, while maintaining a compact model size and moderate training time. This demonstrates that a lightweight, task-specific architecture is capable of capturing salient visual characteristics of mango varieties, such as color distribution, surface texture, and shape, making it suitable for deployment in computationally constrained environments.

Among the models trained from scratch, EfficientNetB0 exhibits improved performance over the CustomCNN, reaching an accuracy of 0.929 and an F1-score of 0.854. ResNet18 trained from scratch achieves the highest overall performance, with an accuracy of 0.966 and a macro F1-score of 0.933. However, analysis of the learning curves indicates that this performance gain comes with increased training instability. While the training loss decreases smoothly, the validation loss shows occasional sharp spikes and corresponding drops in validation accuracy, suggesting sensitivity to sample variability and potential overfitting in later epochs.

Transfer learning provides more stable convergence behavior, particularly for EfficientNetB0. The EfficientNetB0 transfer learning model achieves an accuracy of 0.964 and an F1-score of 0.929, with closely aligned training and validation loss curves throughout training. The smooth and consistent convergence indicates that pretrained representations effectively generalize to mango imagery, reducing training volatility and improving robustness. In contrast, ResNet18 with transfer learning performs slightly worse than its scratch-trained counterpart, suggesting that pretrained features may not fully capture the fine-grained inter-class variations present in mango varieties.

The learning curves show that transfer learning models (EfficientNetB0 and ResNet18) achieve rapid loss reduction within the first few epochs, followed by smooth and monotonic convergence, indicating stable optimization and effective feature reuse. Their training and validation accuracy curves remain closely aligned throughout training, suggesting strong generalization with minimal overfitting. In contrast, scratch-trained EfficientNetB0 exhibits slower early convergence and higher validation loss, reflecting difficulty in learning robust features from random initialization. The ResNet18 trained from scratch attains very high training accuracy but shows abrupt spikes in validation loss and sharp drops in validation accuracy near later epochs, indicating sensitivity to training dynamics and potential overfitting. The CustomCNN demonstrates steady and controlled convergence, with consistently low train–validation gaps, albeit at a slightly lower final accuracy. These observations indicate that while deeper scratch-trained models can reach higher peak accuracy, their learning behavior is less stable. Overall, the curves confirm that transfer learning provides a more reliable and consistent training process on MangoImageBd, whereas CustomCNN offers stable performance with reduced training volatility.
\begin{table*}[t]
\caption{Performance comparison of different models on the MangoImageBd dataset}
\label{tab:performance_table_MangoImageBd}
\centering
\small
\setlength{\tabcolsep}{4pt}
\resizebox{\textwidth}{!}{%
\begin{tabular}{l c c c c c c c}
\toprule
\textbf{Model} 
& \textbf{Acc.} 
& \textbf{Prec.} 
& \textbf{Recall} 
& \textbf{F1} 
& \textbf{Train Time (s)} 
& \textbf{Params} 
& \textbf{Size (MB)} \\
\midrule
CustomCNN 
& 0.907 & 0.829 & 0.845 & 0.821 
& 1932.35 & 1,878,095 & 7.18 \\

EfficientNetB0  
& 0.929 & 0.866 & 0.849 & 0.854 
& 760.47 & 4,026,763 & 15.52 \\

EfficientNetB0 (Transfer) 
& 0.964 & 0.927 & 0.932 & 0.929 
& 1861.47 & 4,026,763 & 15.52 \\

ResNet18 (Scratch) 
& 0.966 & 0.937 & 0.930 & 0.932 
& 608.58 & 11,184,207 & 42.70 \\

ResNet18 (Transfer) 
& 0.945 & 0.882 & 0.887 & 0.883 
& 1827.71 & 11,184,207 & 42.70 \\
\bottomrule
\end{tabular}
}
\end{table*}


\begin{figure}[t]
\centering

\begin{subfigure}{0.49\columnwidth}
  \includegraphics[width=\linewidth]{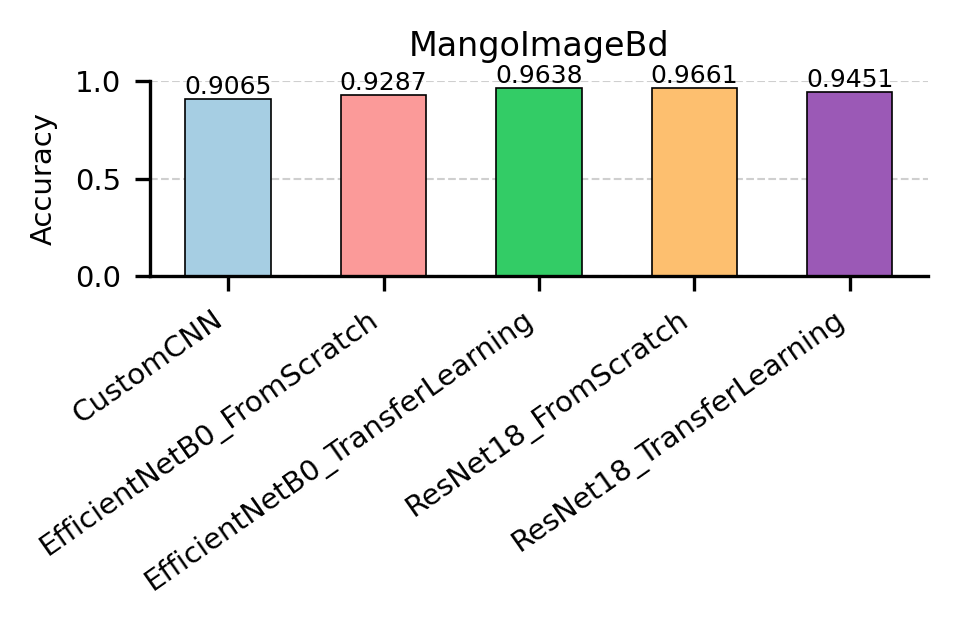}
  \caption{Accuracy}
\end{subfigure}
\hfill
\begin{subfigure}{0.49\columnwidth}
  \includegraphics[width=\linewidth]{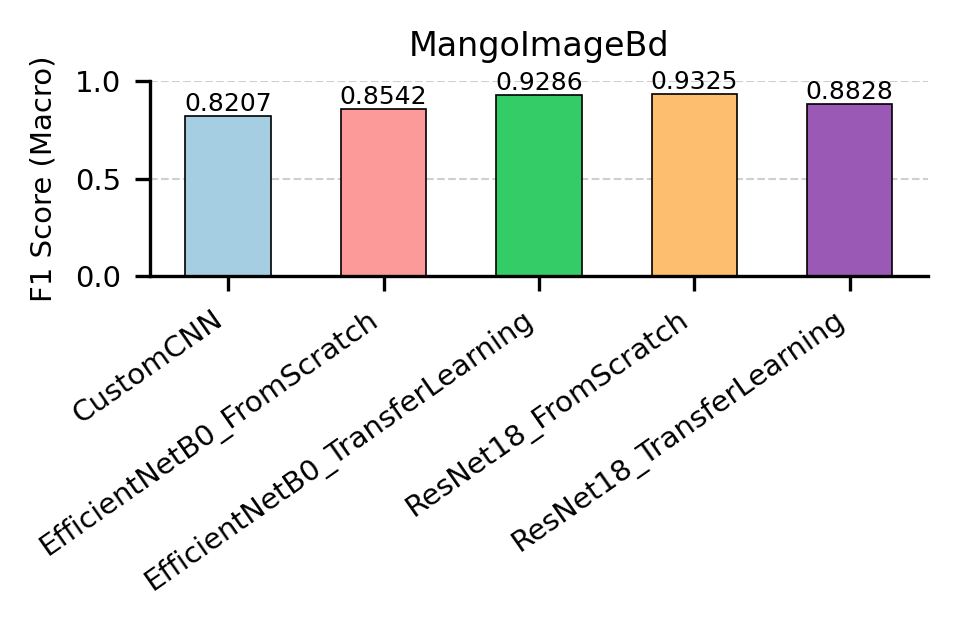}
  \caption{F1-score}
\end{subfigure}

\vspace{1.5mm}

\begin{subfigure}{0.49\columnwidth}
  \includegraphics[width=\linewidth]{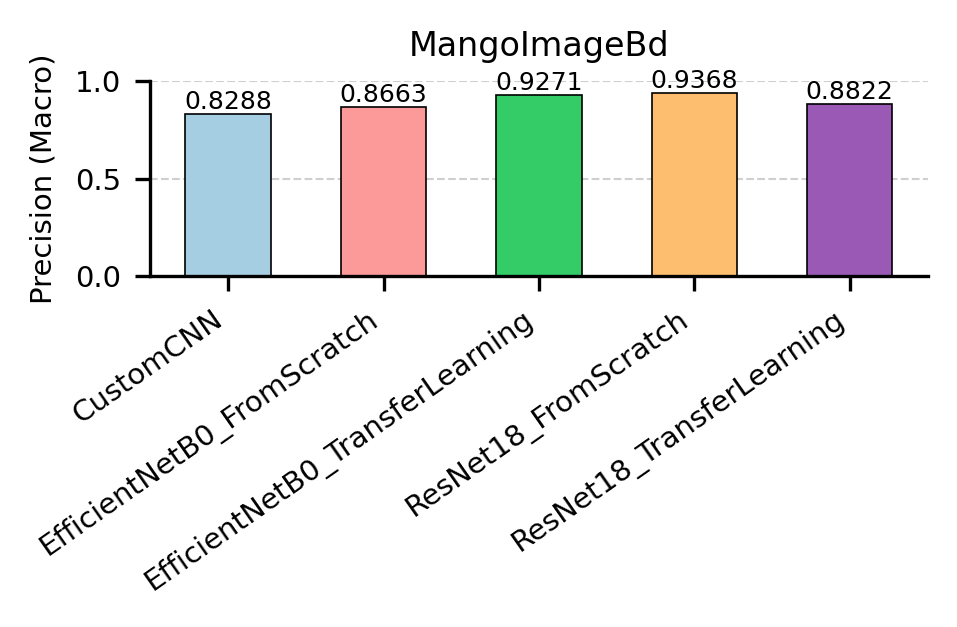}
  \caption{Precision}
\end{subfigure}
\hfill
\begin{subfigure}{0.49\columnwidth}
  \includegraphics[width=\linewidth]{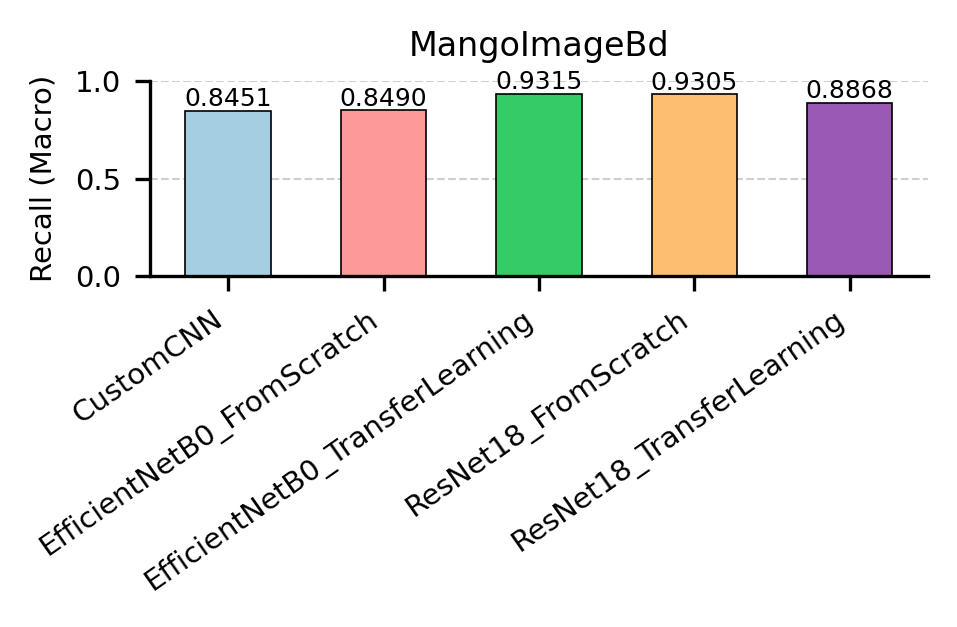}
  \caption{Recall}
\end{subfigure}

\vspace{1.5mm}

\begin{subfigure}{0.49\columnwidth}
  \includegraphics[width=\linewidth]{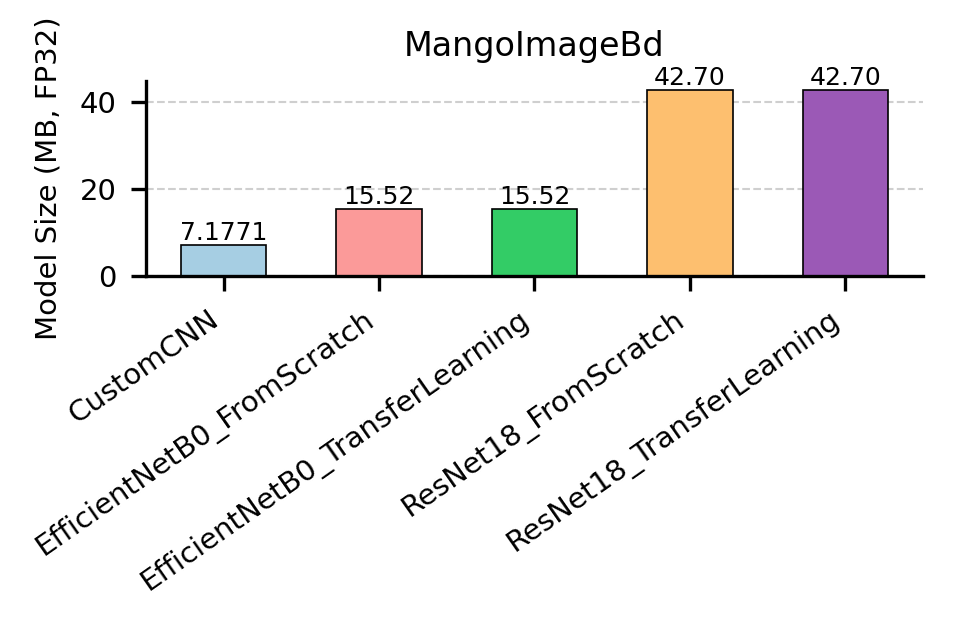}
  \caption{Model size}
\end{subfigure}
\hfill
\begin{subfigure}{0.49\columnwidth}
  \includegraphics[width=\linewidth]{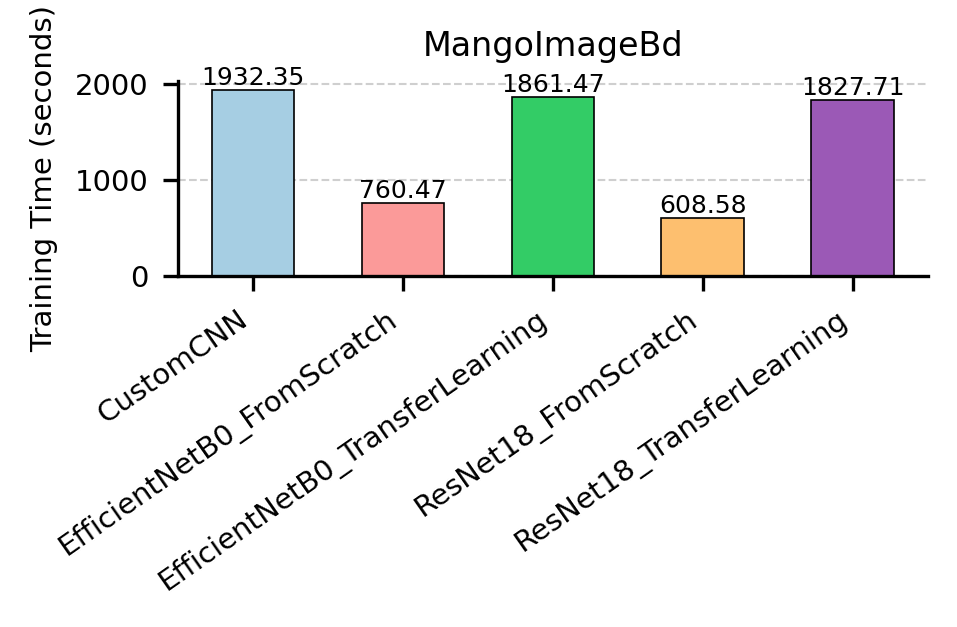}
  \caption{Training time}
\end{subfigure}

\caption{Performance comparison of CustomCNN, scratch-trained, and transfer learning models on the MangoImageBD dataset.}
\label{fig:mango-performance}
\end{figure}

\FloatBarrier

\begin{figure}[htbp]
\centering
\includegraphics[width=0.80\textwidth]{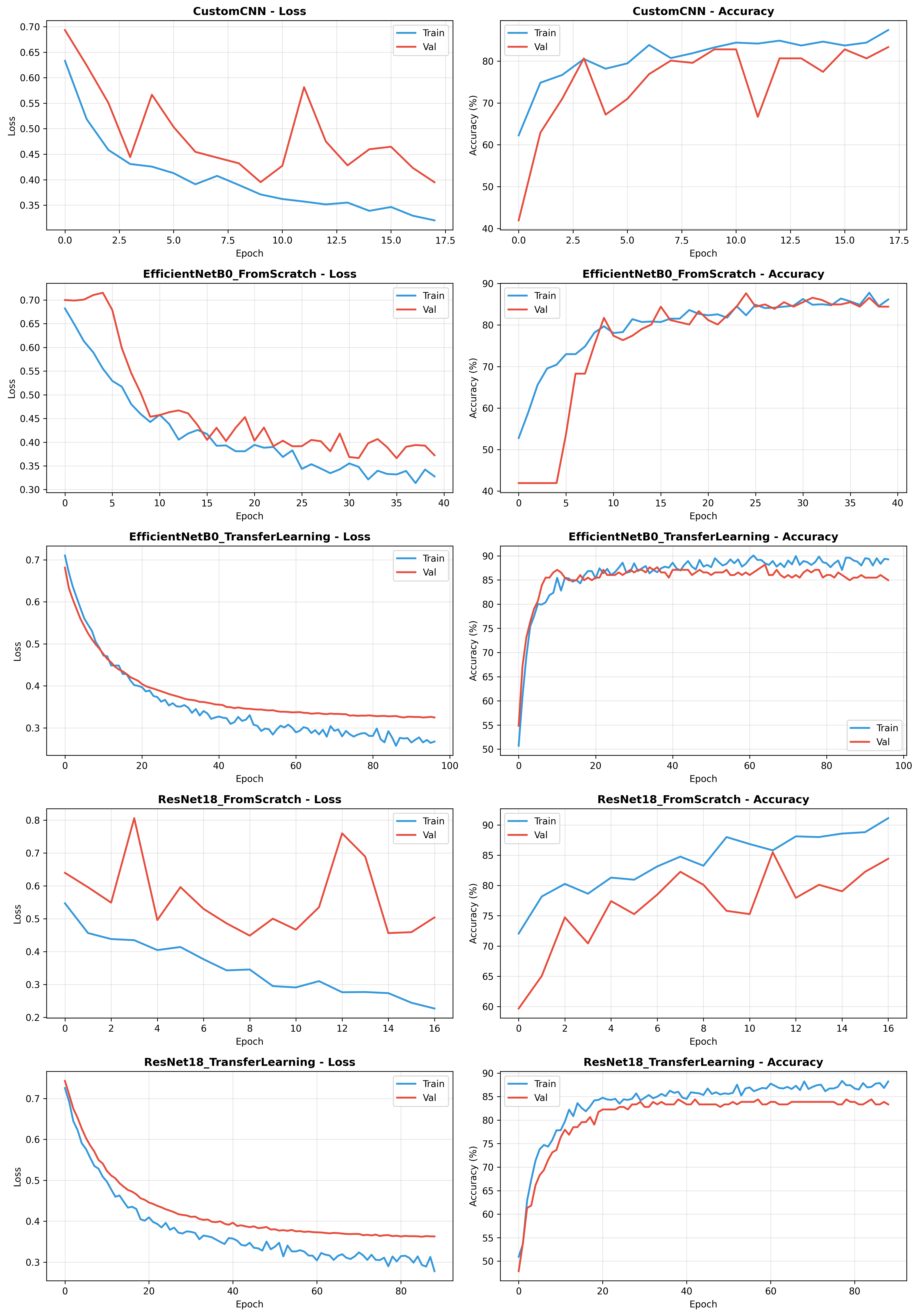} 
\caption{Comaprison of training Curve for testing models on MangoImageBD dataset }
\label{fig:training_curve_mango}
\end{figure}

\textbf{PaddyVariety}\\
The experimental results on the PaddyVariety dataset, summarized in Table~\ref{tab:performance_table_PaddyVariety}, highlight the challenges of fine-grained agricultural image classification and reveal clear differences in convergence behavior across model architectures. The CustomCNN provides a reasonable baseline performance while maintaining a compact architecture and lower computational cost. However, the learning curves indicate noticeable fluctuations in validation loss and accuracy during training, suggesting sensitivity to intra-class similarity and background variations commonly present in paddy variety images.

When trained from scratch, EfficientNetB0 demonstrates improved generalization compared to the CustomCNN, achieving smoother convergence and higher validation accuracy. Nevertheless, the validation loss curve exhibits intermittent spikes during early epochs, indicating instability caused by limited discriminative features between visually similar rice varieties. ResNet18 trained from scratch achieves the strongest overall performance on the PaddyVariety dataset, with consistently increasing validation accuracy and a clear gap over lighter models. Despite this, the learning curves reveal significant oscillations in validation loss during early training stages, pointing to overfitting tendencies and sensitivity to sample imbalance.

Transfer learning shows mixed effectiveness on the PaddyVariety dataset. Although both EfficientNetB0 and ResNet18 with pretrained weights exhibit smoother loss decay and more stable convergence compared to their scratch-trained counterparts, their final classification performance is lower. This suggests that features learned from large-scale natural image datasets may not adequately capture the subtle visual differences required for distinguishing paddy varieties. The relatively small gap between training and validation curves indicates stable optimization, but limited feature transferability restricts overall performance gains.

The learning curves show that PaddyVarietyBD presents a challenging fine-grained classification scenario, as reflected by pronounced fluctuations in validation loss and accuracy across several models. The CustomCNN exhibits stable but relatively slow convergence, with noticeable oscillations in validation accuracy during early and mid-training epochs, indicating limited discriminative capacity for subtle inter-class variations. EfficientNetB0 trained from scratch converges gradually, but its validation curves display repeated spikes in loss and accuracy, suggesting sensitivity to data variability. Transfer learning with EfficientNetB0 improves convergence smoothness and reduces loss volatility, although its validation accuracy saturates at a lower level. In contrast, ResNet18 trained from scratch demonstrates a consistent reduction in training loss and a steady rise in validation accuracy, despite early instability, ultimately achieving the highest sustained accuracy. The transfer-learning variant of ResNet18 shows smoother convergence but slightly lower validation accuracy. These trends indicate that learning domain-specific features from scratch is particularly beneficial for the fine-grained nature of PaddyVarietyBD.
\begin{table*}[t]
\caption{Performance comparison of different models on the PaddyVariety dataset}
\label{tab:performance_table_PaddyVariety}
\centering
\small
\setlength{\tabcolsep}{4pt}
\resizebox{\textwidth}{!}{%
\begin{tabular}{l c c c c c c c}
\toprule
\textbf{Model} 
& \textbf{Acc.} 
& \textbf{Prec.} 
& \textbf{Recall} 
& \textbf{F1} 
& \textbf{Train Time (s)} 
& \textbf{Params} 
& \textbf{Size (MB)} \\
\midrule
CustomCNN 
& 0.788 & 0.803 & 0.788 & 0.783 
& 8337.61 & 1,888,355 & 7.22 \\

EfficientNetB0 
& 0.866 & 0.871 & 0.866 & 0.865 
& 5773.28 & 4,052,383 & 15.62 \\

EfficientNetB0 (Transfer) 
& 0.664 & 0.674 & 0.664 & 0.653 
& 6233.85 & 4,052,383 & 15.62 \\

ResNet18 (Scratch) 
& 0.917 & 0.918 & 0.917 & 0.916 
& 6138.50 & 11,194,467 & 42.74 \\

ResNet18 (Transfer) 
& 0.637 & 0.641 & 0.637 & 0.627 
& 6074.48 & 11,194,467 & 42.74 \\
\bottomrule
\end{tabular}
}
\end{table*}

\begin{figure}[t]
\centering

\begin{subfigure}{0.49\columnwidth}
  \includegraphics[width=\linewidth]{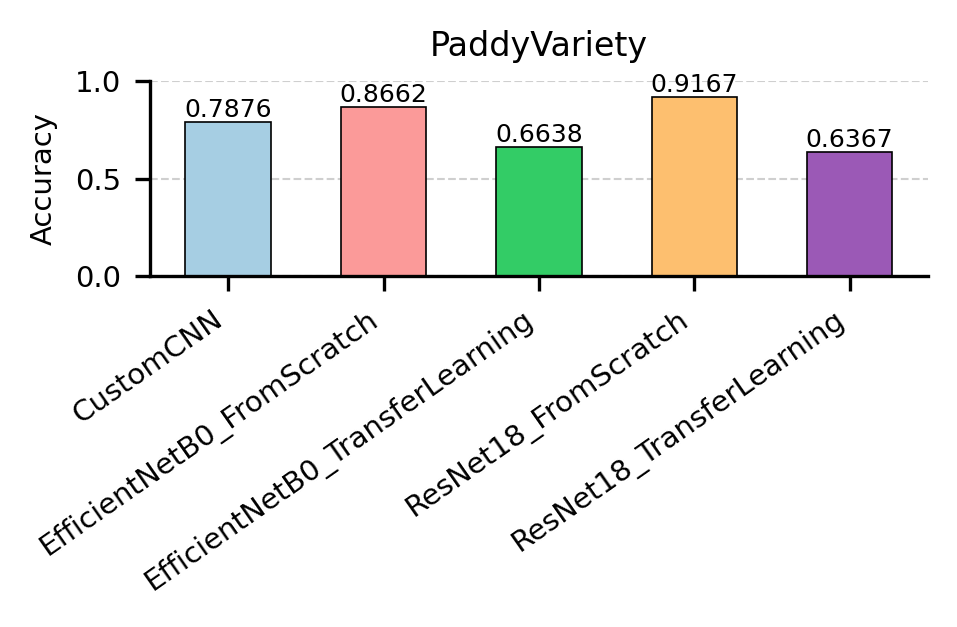}
  \caption{Accuracy}
\end{subfigure}
\hfill
\begin{subfigure}{0.49\columnwidth}
  \includegraphics[width=\linewidth]{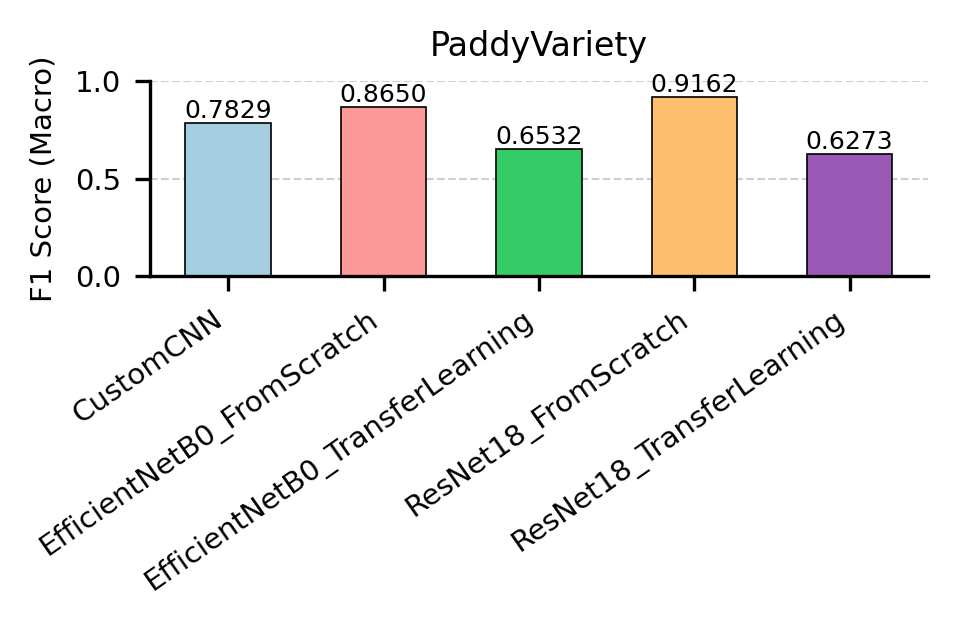}
  \caption{F1-score}
\end{subfigure}

\vspace{1.5mm}

\begin{subfigure}{0.49\columnwidth}
  \includegraphics[width=\linewidth]{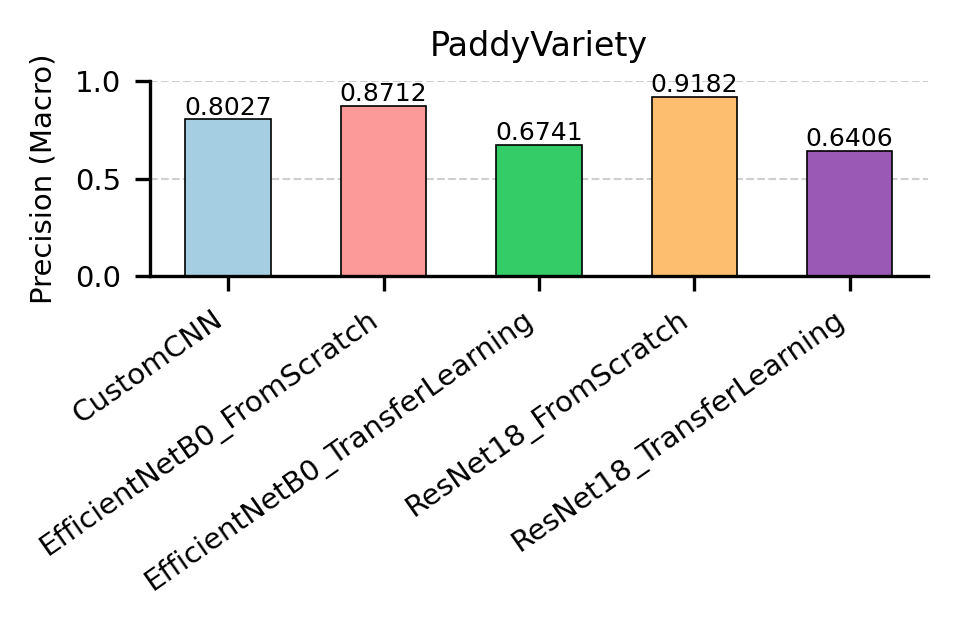}
  \caption{Precision}
\end{subfigure}
\hfill
\begin{subfigure}{0.49\columnwidth}
  \includegraphics[width=\linewidth]{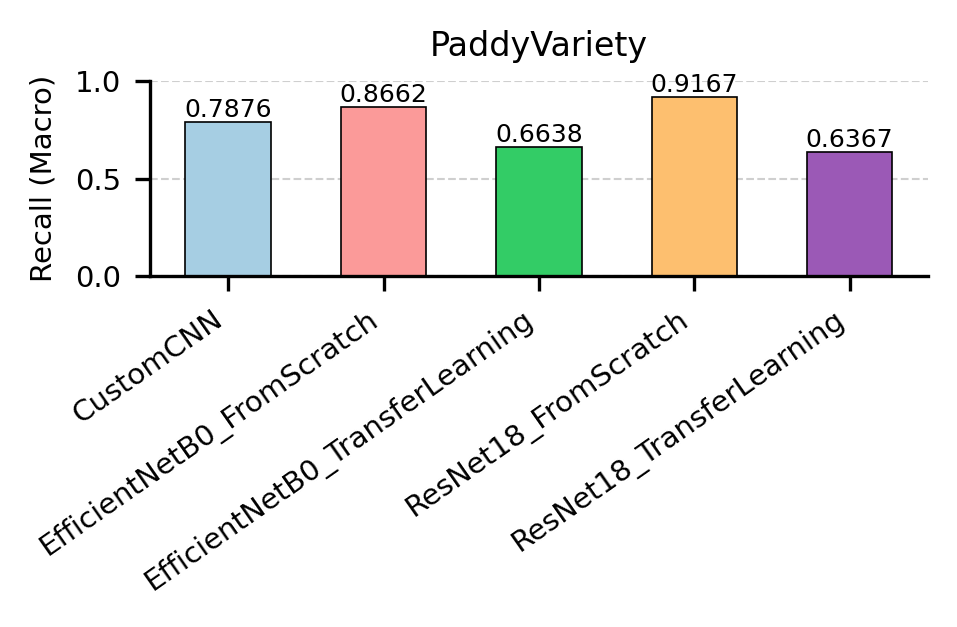}
  \caption{Recall}
\end{subfigure}

\vspace{1.5mm}

\begin{subfigure}{0.49\columnwidth}
  \includegraphics[width=\linewidth]{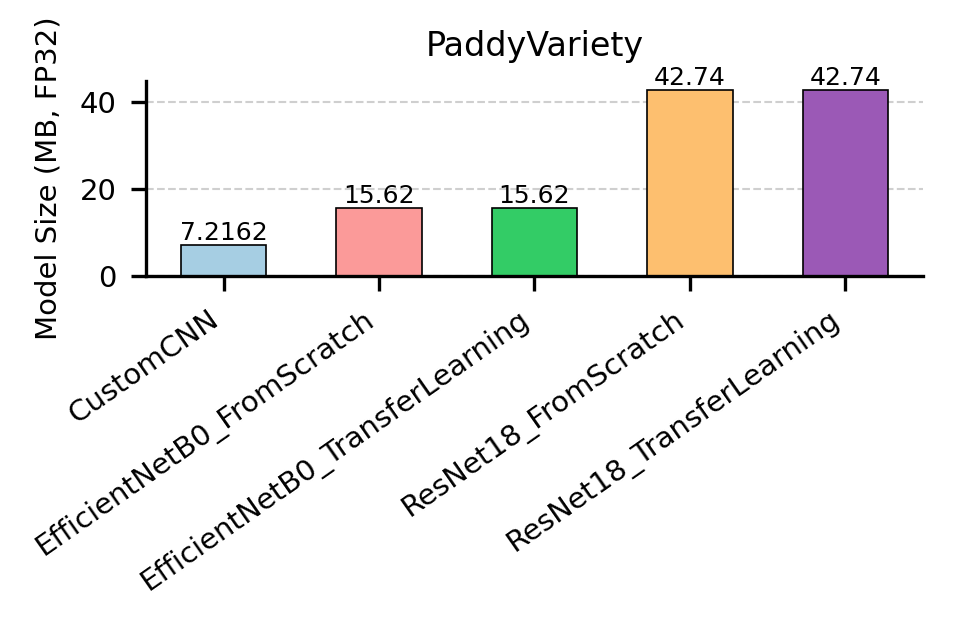}
  \caption{Model size}
\end{subfigure}
\hfill
\begin{subfigure}{0.49\columnwidth}
  \includegraphics[width=\linewidth]{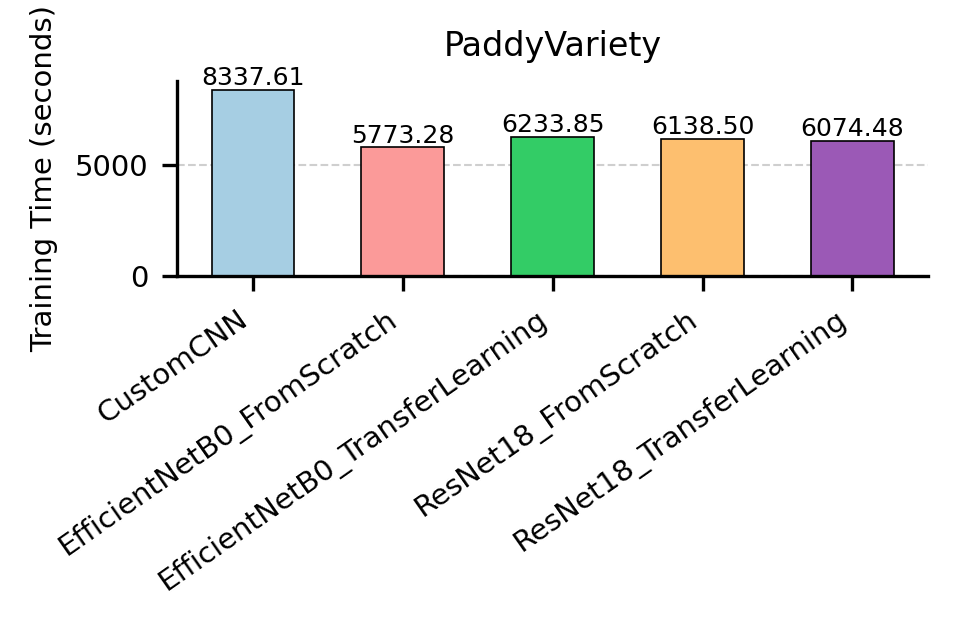}
  \caption{Training time}
\end{subfigure}

\caption{Performance comparison of CustomCNN, scratch-trained, and transfer learning models on the PaddyVarietyBD dataset.}
\label{fig:performance_compare_paddy}
\end{figure}

\FloatBarrier

\begin{figure}[htbp]
\centering
\includegraphics[width=0.80\textwidth]{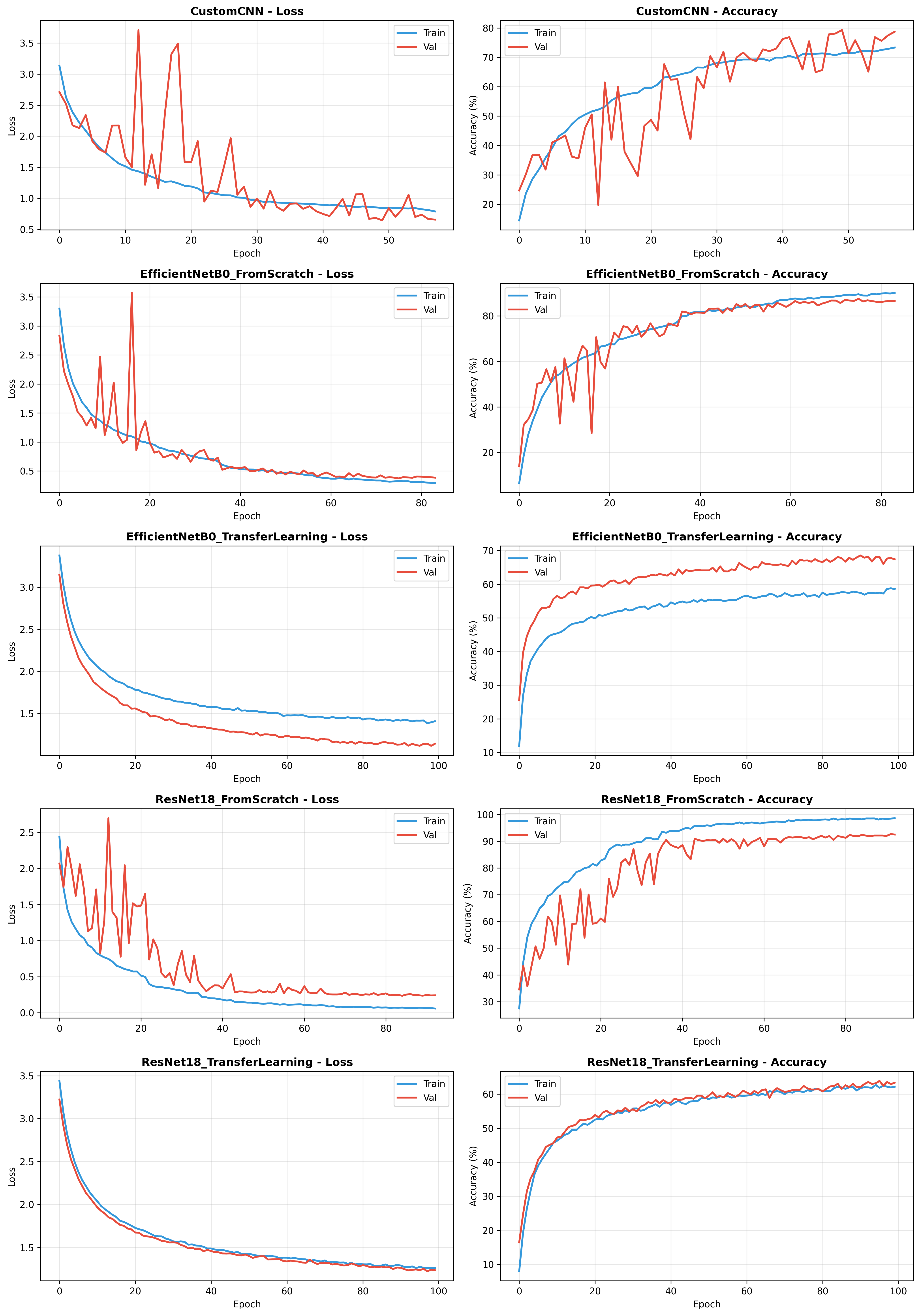} 
\caption{Comaprison of training Curve for testing models on PaddyVarietyBD dataset }
\label{fig:training_curve_paddy}
\end{figure}

\textbf{Auto-RickshawImageBD}\\
The experimental results on the Auto-RickshawImageBD dataset, summarized in Table~\ref{tab:performance_table_AutoRickshawImageBD}, demonstrate the effectiveness of different learning strategies for binary vehicle image classification, with clear distinctions in convergence behavior and generalization performance across model architectures. The CustomCNN achieves a reasonable baseline performance, with an accuracy of 0.720 and a macro F1-score of 0.678, while maintaining a compact architecture and reduced memory footprint. However, the learning curves reveal noticeable fluctuations in validation loss and accuracy, indicating sensitivity to background clutter, illumination changes, and viewpoint variations commonly present in real-world auto-rickshaw imagery.

When trained from scratch, EfficientNetB0 exhibits unstable learning behavior, achieving relatively low accuracy (0.600) and macro F1-score (0.454). The corresponding learning curves show sharp oscillations in validation performance, suggesting difficulty in learning discriminative features from random initialization under limited data conditions. In contrast, ResNet18 trained from scratch demonstrates stronger representational capacity, achieving an accuracy of 0.805 and a macro F1-score of 0.723, although early-stage validation loss spikes indicate potential overfitting and sensitivity to sample imbalance.

Transfer learning substantially improves convergence stability and generalization. EfficientNetB0 with transfer learning achieves an accuracy of 0.744 and a macro F1-score of 0.704, with smooth loss decay and closely aligned training and validation accuracy curves. ResNet18 with transfer learning shows comparable behavior, achieving an accuracy of 0.740 and a macro F1-score of 0.701. The reduced gap between training and validation curves confirms that pretrained representations effectively capture structural cues such as vehicle shape and edge patterns. Overall, the learning curve analysis confirms that transfer learning provides the most reliable performance on Auto-RickshawImageBD, while CustomCNN remains a viable lightweight alternative when computational efficiency is prioritized.\\
The learning curves indicate that the AutoRickshawImageBD task exhibits notable training instability, particularly during early epochs, reflecting variability in the dataset. The CustomCNN shows gradual loss reduction but frequent oscillations in validation accuracy, indicating sensitivity to sample-level variations. EfficientNetB0 trained from scratch displays sharp validation accuracy spikes and relatively flat training trends, suggesting difficulty in learning robust features from random initialization. Transfer learning with EfficientNetB0 improves convergence smoothness and stabilizes validation performance. ResNet18 trained from scratch achieves high training accuracy but exhibits pronounced validation loss spikes, indicating overfitting tendencies. In contrast, ResNet18 with transfer learning demonstrates stable loss convergence and closely aligned training–validation accuracy curves. Overall, transfer learning provides the most consistent and reliable training behavior for AutoRickshawImageBD.
\begin{table*}[t]
\caption{Performance comparison of different models on the Auto-RickshawImageBD dataset}
\label{tab:performance_table_AutoRickshawImageBD}
\centering
\small
\setlength{\tabcolsep}{4pt}
\resizebox{\textwidth}{!}{%
\begin{tabular}{l c c c c c c c}
\toprule
\textbf{Model} 
& \textbf{Acc.} 
& \textbf{Prec.} 
& \textbf{Recall} 
& \textbf{F1} 
& \textbf{Train Time (s)} 
& \textbf{Params} 
& \textbf{Size (MB)} \\
\midrule
CustomCNN 
& 0.720 & 0.676 & 0.736 & 0.678 
& 2239.49 & 1,871,426 & 7.15 \\

EfficientNetB0
& 0.600 & 0.589 & 0.600 & 0.454 
& 874.31 & 4,010,110 & 15.46 \\

EfficientNetB0 (Transfer) 
& 0.744 & 0.693 & 0.744 & 0.704 
& 6838.75 & 4,010,110 & 15.46 \\

ResNet18 (Scratch) 
& 0.805 & 0.728 & 0.718 & 0.723 
& 1178.32 & 11,177,538 & 42.68 \\

ResNet18 (Transfer) 
& 0.740 & 0.697 & 0.764 & 0.701 
& 5328.38 & 11,177,538 & 42.68 \\
\bottomrule
\end{tabular}
}
\end{table*}

\begin{figure}[t]
\centering

\begin{subfigure}{0.49\columnwidth}
  \includegraphics[width=\linewidth]{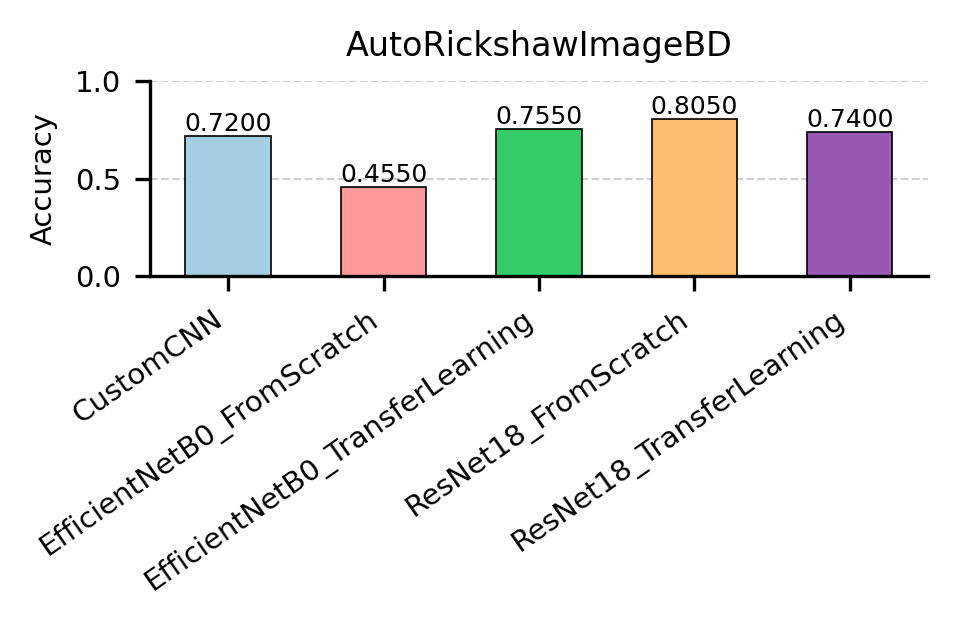}
  \caption{Accuracy}
\end{subfigure}
\hfill
\begin{subfigure}{0.49\columnwidth}
  \includegraphics[width=\linewidth]{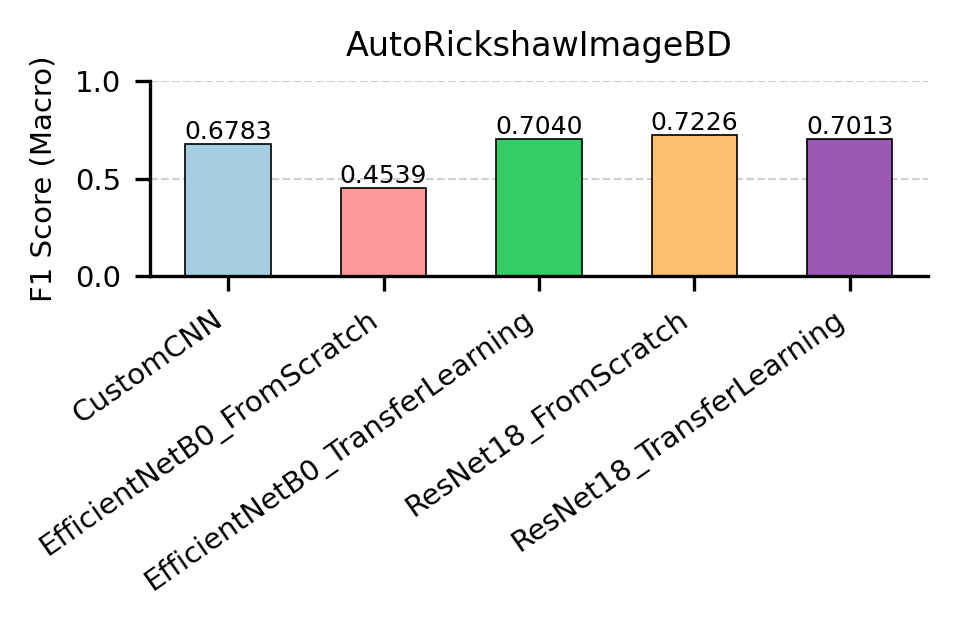}
  \caption{F1-score}
\end{subfigure}

\vspace{1.5mm}

\begin{subfigure}{0.49\columnwidth}
  \includegraphics[width=\linewidth]{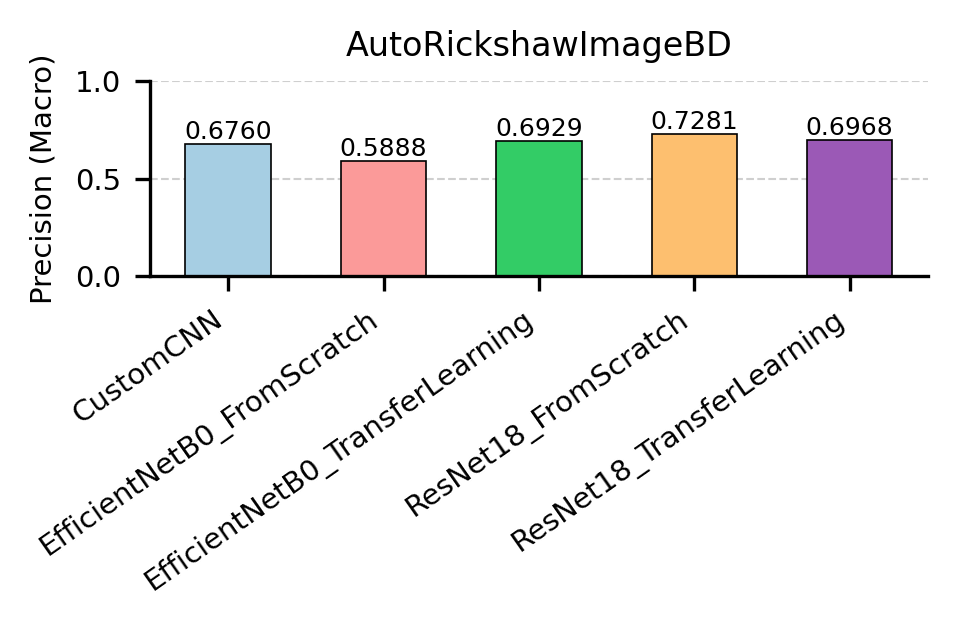}
  \caption{Precision}
\end{subfigure}
\hfill
\begin{subfigure}{0.49\columnwidth}
  \includegraphics[width=\linewidth]{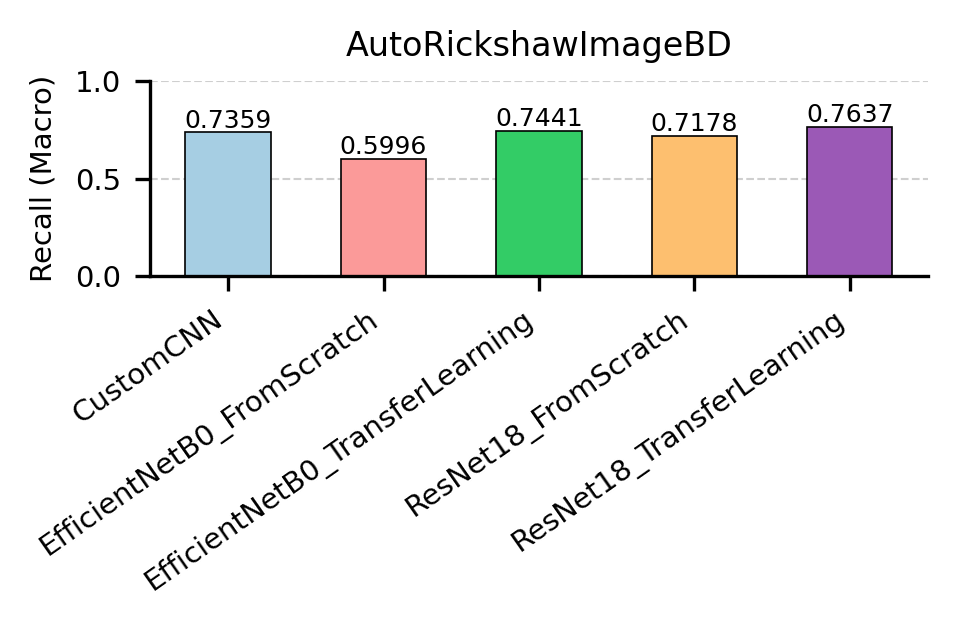}
  \caption{Recall}
\end{subfigure}

\vspace{1.5mm}

\begin{subfigure}{0.49\columnwidth}
  \includegraphics[width=\linewidth]{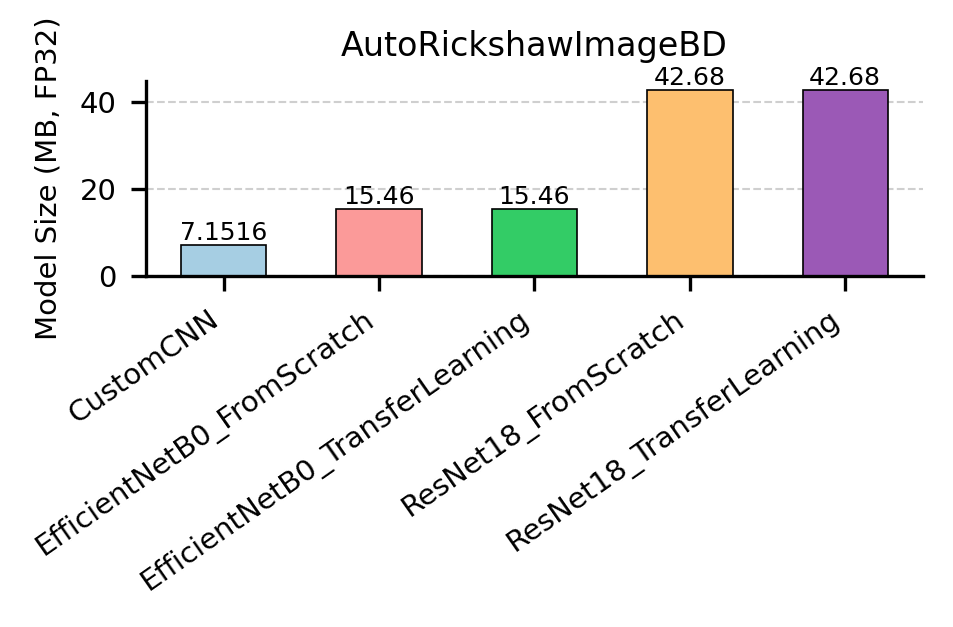}
  \caption{Model size}
\end{subfigure}
\hfill
\begin{subfigure}{0.49\columnwidth}
  \includegraphics[width=\linewidth]{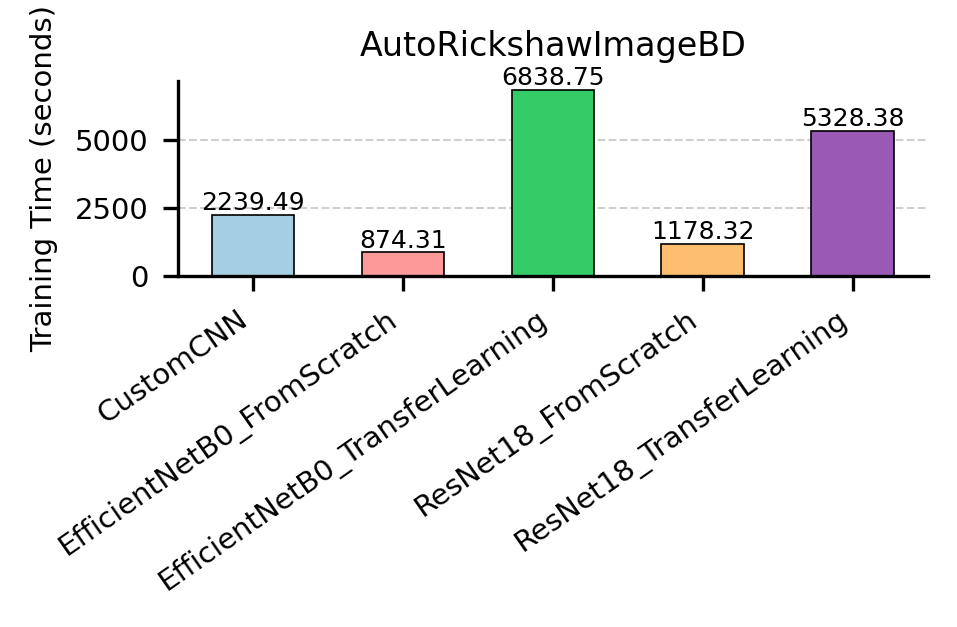}
  \caption{Training time}
\end{subfigure}

\caption{Performance comparison of CustomCNN, scratch-trained, and transfer learning models on the AutoRickshawImageBD dataset.}
\label{fig:performance_compare_auto}
\end{figure}

\FloatBarrier

\begin{figure}[htbp]
\centering
\includegraphics[width=0.80\textwidth]{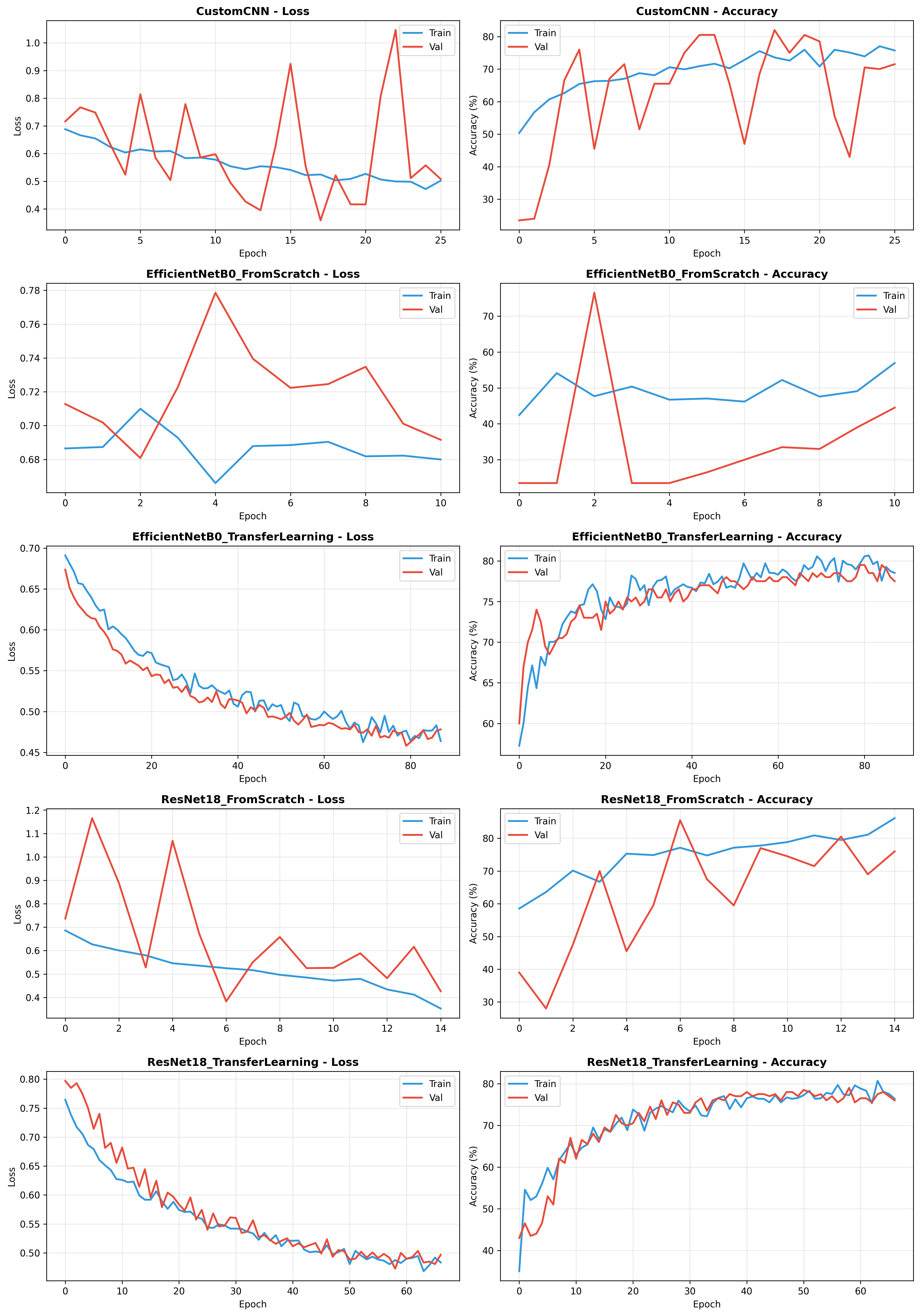} 
\caption{Comaprison of training Curve for testing models on AutoRickshawImageBD dataset}
\label{fig:training_curve_auto}
\end{figure}

\section{Discussion}
The experimental results reveal clear trade-offs between classification performance, model complexity, and computational efficiency across the evaluated architectures. Lightweight, task-specific models such as the CustomCNN consistently demonstrate low memory requirements and reduced computational cost, making them suitable for deployment in resource-constrained environments. However, their limited representational capacity results in lower accuracy and macro F1-scores when compared to deeper architectures, particularly on datasets with complex visual patterns or fine-grained class distinctions.

In contrast, deeper models such as EfficientNetB0 and ResNet18 generally achieve superior classification performance due to their increased depth and expressive power. These gains are more pronounced when transfer learning is applied, as pretrained representations provide improved initialization, faster convergence, and enhanced generalization. Nevertheless, the improved accuracy achieved by these models comes at the expense of higher parameter counts, increased memory footprints, and longer training times, which may limit their applicability in real-time or embedded deployment scenarios.

The comparison between training from scratch and transfer learning further highlights application-dependent behavior. While transfer learning improves training stability and reduces overfitting in most cases, it does not universally yield the highest performance, particularly for fine-grained or domain-specific datasets where pretrained features may not optimally align with the target task. In such cases, scratch-trained deep models can achieve competitive or superior results, albeit with increased sensitivity to training dynamics and data distribution.

Overall, the choice of model should be guided by both performance requirements and resource constraints. The CustomCNN is best suited for applications prioritizing efficiency, compactness, and ease of deployment. EfficientNetB0 with transfer learning offers a balanced trade-off between accuracy and computational cost, making it a practical choice for most real-world scenarios. ResNet18, particularly when trained from scratch, is preferable when maximizing predictive performance is the primary objective and sufficient computational resources are available. These findings emphasize that no single architecture is universally optimal, and model selection should be driven by the specific characteristics and constraints of the target application.


\section*{Conclusion}
This study evaluated custom-designed and deep convolutional neural network architectures across multiple domain-specific image classification tasks. The results indicate that lightweight models offer strong computational efficiency and deployment feasibility, whereas deeper architectures achieve higher accuracy at the cost of increased complexity. Transfer learning improves training stability and generalization when data are limited, while scratch-trained models can be more effective for fine-grained or highly domain-specific tasks.

Overall, model selection should be guided by accuracy requirements and computational constraints: EfficientNetB0 with transfer learning provides a balanced solution, CustomCNN suits resource-constrained environments, and deeper scratch-trained models are preferable when maximizing predictive performance is the primary objective.

\clearpage
\bibliographystyle{plain}
\bibliography{sn-bibliography}

\end{document}